\documentclass[conference]{IEEEtran}
\IEEEoverridecommandlockouts
\usepackage[T1]{fontenc}
\usepackage{graphicx}
\usepackage{booktabs}
\usepackage{amsmath}
\usepackage{array}
\usepackage{makecell}
\usepackage{multirow}
\usepackage{url}
\usepackage{cite}
\usepackage{stfloats}
\usepackage{balance}
\usepackage{placeins}
\usepackage{float}
\usepackage{capt-of}
\usepackage{xcolor}
\usepackage{comment}
\usepackage{microtype}
\usepackage{amsmath,amssymb}

\newcolumntype{L}[1]{>{\raggedright\arraybackslash}p{#1}}
\newcolumntype{C}[1]{>{\centering\arraybackslash}p{#1}}
\title{NEXUS-MI: Communication-Aware Federated Personalization for Gateway-Coordinated Motor-Imagery Brain-Computer Interfaces}

\author{
    Daniel Adu Worae\\
    University of Notre Dame\\
    \texttt{dworae@nd.edu}
    \and
    Aarthy Nagarajan\\
    University of Notre Dame\\
    \texttt{anagaraj@nd.edu}

\thanks{Daniel Adu Worae is with the Department of Computer Science and Engineering, University of Notre Dame, 307 Fitzpatrick Hall, Notre Dame, IN 46556, USA (e-mail: dworae@nd.edu).}%
\thanks{Aarthy Nagarajan is with the Department of Computer Science and Engineering and the Lucy Family Institute for Data and Society, University of Notre Dame, Notre Dame, IN 46556, USA (e-mail: anagaraj@nd.edu).}%
\thanks{Daniel Adu Worae and Aarthy Nagarajan are corresponding authors.}%
}
\begin{document}
\maketitle

\begin{abstract}
Electroencephalography (EEG)-based motor-imagery brain-computer interfaces (MI-BCIs) exhibit variability across subjects and sessions, complicating personalization from limited calibration data. Federated learning can exploit shared representations without centralizing raw EEG. However, existing federated MI studies assume regular synchronization and overlook intermittent gateway availability and coordination. We introduce \textit{NEXUS-MI}, a gateway-coordinated federated personalization framework that formulates synchronization as a coupled learning-and-communication control problem. During the gateway-coordinated phase, raw EEG and classifier heads remain local while an edge coordinator maintains the shared backbone. We evaluate \textit{NEXUS-MI} through offline replay using BCI Competition IV Dataset 2a (BCICIV-2a; 9 subjects, 4 classes) and OpenBMI (54 subjects, 2 classes). Session~1 supports backbone learning, and Session~2 provides limited-calibration personalization and held-out testing. An ideal-link reference and six heterogeneous-link policies characterize gateway participation, buffering, stale-update admission, and backbone-download control. The principal comparison holds delayed-update handling fixed while contrasting non-adaptive and communication-aware synchronization. Paired subject-level comparisons use Holm adjustment, and robustness across five matched realizations is assessed by hierarchical bootstrap. Communication-aware coordination reduced server-to-client backbone traffic by approximately \(42\%\) on both datasets, while cohort-level accuracy differences were small and realization-dependent. Cohort averages also concealed subject-level vulnerability, with losses reaching approximately \(12\) percentage points on BCICIV-2a relative to the ideal-link reference. These findings establish gateway synchronization as an explicit design variable in federated MI personalization and motivate joint evaluation of personalized accuracy, communication cost, update freshness, and subject-level reliability.
\end{abstract}

\section{Introduction}
Motor-imagery brain-computer interfaces (MI-BCIs) offer a promising path toward neurorehabilitation systems that can translate imagined movement into control signals without requiring overt motor execution \cite{cervera2018brain,singh2021comprehensive}. This capability is especially important for longitudinal rehabilitation, where users may need repeated practice beyond specialized clinical facilities \cite{arpaia2023paving}. However, the transition from controlled MI-BCI experiments to reliable home use remains limited by decoder stability. EEG-based MI signals are weak, non-stationary, and highly user-dependent, with substantial variation across subjects, sessions, and recording conditions \cite{cho2017eeg,wu2020transfer}. Decoders are affected by subject physiology, electrode placement, attention, fatigue, and session-to-session drift \cite{ahn2015performance,myrden2015effects,xiong2025adaptive}. A model that performs well during calibration may degrade when used by the same subject in a later session or transferred to another user. In MI-BCI rehabilitation, decoder instability has consequences beyond reduced classification performance: inconsistent predictions can weaken the correspondence between motor imagery and therapeutic feedback, disrupt feedback-driven practice, and increase the need for supervised recalibration. This risk is especially important for users with low or unstable MI-BCI control, for whom further degradation may make the resulting feedback too unreliable to support continued use \cite{ahn2015performance,lee2019eeg}. Maintaining dependable decoder performance for each user across sessions in deployment-oriented MI-BCI systems is, therefore, a central requirement for sustained rehabilitation use.

Existing adaptation strategies address this reliability challenge only partially. Subject-specific training can adapt to individual EEG structure; however, in routine home use, repeatedly collecting large labeled calibration sets is impractical. Training a deep decoder from small calibration sets risks overfitting, while lightweight adaptation may be limited when the representation being adapted is weak or stale. Centralized multi-user training can exploit population-level EEG structure, but it requires moving sensitive neurophysiological recordings away from the user environment. Federated learning (FL) offers a natural alternative: multiple users can contribute to improving a shared model while keeping raw EEG data local. Prior MI-BCI FL studies have shown that collaborative learning can improve motor-imagery classification and address cross-user heterogeneity without centralizing EEG recordings \cite{jia2024federated,ju2020federated,liu2024aggregating}. These studies establish an important foundation for distributed MI learning, but they largely evaluate FL as a modeling protocol rather than as a gateway-coordinated deployment workflow.

This distinction matters because home MI-BCI is also an edge coordination problem. A target deployment model includes a wearable sensor, a subject-side gateway, and an edge coordinator. The gateway performs local inference and local personalization close to the user, while the edge coordinator maintains a shared representation across users. In this setting, the gateway-to-edge link is not a passive transport channel. Gateways may be intermittently online, bandwidth may vary across homes, updates may be delayed, and local personalization may continue from a stale backbone while a gateway waits for a refreshed model. These events influence which users contribute to the shared backbone, how fresh their updates are when they arrive, which model version each gateway uses locally, and how quickly improved representations reach the user. Consequently, deployment-oriented MI-BCI evaluation must consider not only whether collaborative learning improves decoding, but also whether the personalized decoder available to each subject remains reliable when gateway participation, update delivery, and backbone synchronization are constrained.

Current MI-BCI FL evaluations do not sufficiently expose this deployment dimension. They typically show that federation can support distributed EEG model training, but they do not treat gateway participation, buffering, stale-update admission, and backbone download control as first-order factors in personalized decoding. As a result, the accuracy-communication tradeoff remains under-specified for home rehabilitation, where limited calibration, session drift, subject-level variability, and heterogeneous connectivity can jointly determine whether personalized decoding performance is preserved after deployment. The gap is therefore specific: MI-BCI FL needs a deployment-centered evaluation in which communication policy is treated as a determinant of personalized decoding performance, not only as a background systems cost.

Accordingly, this study examines how gateway synchronization shapes decoding accuracy, communication cost, update freshness, and subject-level reliability in federated MI-BCI personalization under limited calibration and heterogeneous connectivity. Unlike prior MI-BCI FL evaluations that largely report classification performance under regular federation, NEXUS-MI is positioned not as a general-purpose FL optimizer, but as a deployment-centered framework that makes synchronization behavior experimentally testable by relating gateway participation, buffering, delayed-update admission, and backbone-download control to the personalized decoder ultimately available to each subject, with outcomes evaluated against an ideal-link reference.

The study is organized around three research questions.

\textbf{RQ1: How well does gateway-coordinated personalization perform under ideal synchronization?}
We first establish an ideal-link reference for SB-PH and EIB-PH. This condition establishes a reference learning and synchronization trajectory under unconstrained gateway-to-edge communication and provides the baseline needed to interpret communication-constrained operation.

\textbf{RQ2: How do communication policies change the accuracy-communication tradeoff?}
We evaluate gateway-to-edge policies that vary participation, buffering, stale-update admission, communication-aware scheduling, and version-aware backbone download control. This analysis determines whether the communication policy affects only the synchronization cost or also changes the decoding performance achieved through collaborative personalization.

\textbf{RQ3: Which policies preserve subject-level reliability under heterogeneous links?}
We examine whether aggregate accuracy hides uneven effects across users by measuring worst-subject performance, threshold violations, per-subject deviations from the ideal-link reference, and sensitivity to link availability.

We evaluate the framework on two public MI datasets with complementary characteristics: BCICIV-2a, a four-class benchmark with nine subjects and two sessions, and OpenBMI, a larger two-class benchmark with 54 subjects and two sessions \cite{tangermann2012review,lee2019eeg}. The session structure supports a deployment-oriented protocol: earlier session data support shared-backbone learning, while later session data emulate continued use with limited calibration and held-out testing. Across datasets and personalization regimes, we compare an ideal-link reference with six heterogeneous-link communication policies. The evaluation reports on accuracy, client-to-server traffic, server-to-client backbone synchronization, accepted-update staleness, dropped or rejected updates, avoided downloads, link-availability sensitivity, and subject-level reliability.

The results show that gateway communication policy materially changes the operating point of federated MI-BCI personalization. In the strongest observed setting, P5 increased mean OpenBMI EIB-PH accuracy by 0.99 percentage points relative to matched P3 while reducing server-to-client backbone traffic by approximately 41\%. However, accuracy effects varied across datasets and personalization regimes, with BCICIV-2a exhibiting greater subject-level sensitivity under constrained communication. Across five matched training and gateway-availability realizations, the server-to-client traffic reduction remained approximately 42\% on both datasets, whereas the P5--P3 accuracy differences were small, varied in direction, and had crossed hierarchical-bootstrap 95\% confidence intervals that included zero. These findings show that deployment-oriented MI-BCI evaluation should account for synchronization costs, update freshness, and reliability across users, rather than mean decoding accuracy alone.

This paper makes the following contributions:

\begin{itemize}
\item \textbf{Gateway-coordinated MI-BCI personalization framework.}
We introduce an edge-enabled MI-BCI framework that separates subject-local classifier-head adaptation from cohort-level backbone learning, enabling limited-calibration personalization while retaining raw EEG and subject-specific heads at the gateway during the gateway-coordinated phase.

\item \textbf{Communication-policy formulation for MI-BCI FL.}
We formulate gateway synchronization as a configurable policy space spanning gateway participation, buffering, stale-update admission, and version-aware backbone download control. This formulation makes the coordination mechanisms governing shared-backbone exchange explicit design variables rather than implicit implementation assumptions.

\item \textbf{Controlled deployment-oriented evaluation.}
We evaluate NEXUS-MI on BCICIV-2a and OpenBMI using a session-based protocol that contrasts an ideal-link reference with six policies under heterogeneous gateway availability and matched personalization regimes. This controlled comparison quantifies how synchronization policy affects personalized decoding, communication traffic, update freshness, delayed-update rejection, and avoided backbone downloads.

\item \textbf{Subject-level reliability analysis.}
We show how aggregate accuracy can obscure uneven subject-level effects under constrained communication by measuring worst-subject performance, threshold violations, per-subject deviation from the ideal-link reference, and link-availability sensitivity. This exposes whether communication-efficient policies preserve reliable decoding across users.

\end{itemize}

The rest of the paper is organized as follows. Section~II reviews related work. Section~III presents the system model and problem formulation. Section~IV describes the gateway-coordinated federated personalization framework. Section~V presents the experimental evaluation. Section~VI discusses deployment implications. Section~VII concludes the paper.

\section{Related Work}

\subsection{MI-BCI Decoding, Calibration, and Personalization}

Motor-imagery EEG decoding has a long history of methods that seek discriminative sensorimotor-rhythm structure from noisy, low-SNR signals. Classical pipelines based on common spatial patterns and filter-bank common spatial patterns remain influential because they combine spatial filtering with frequency-specific MI information \cite{ang2012filter}. Deep EEG models extend this tradition by learning temporal, spectral, and spatial representations directly from data. Shallow and deep convolutional EEG decoders, compact architectures such as EEGNet, and MI-focused models such as FBCNet have established strong neural backbones for MI classification under limited-sample conditions \cite{schirrmeister2017deep,lawhern2018eegnet,mane2021fbcnet}. More recent work has continued this progression by integrating established spatial priors with end-to-end representation learning; CSP-Net, for example, incorporates common-spatial-pattern structure into neural models for both subject-dependent and subject-independent MI decoding \cite{jiang2024csp}. These works improve the representational substrate for MI decoding, but they do not address how such representations should be maintained across users when the deployment system is distributed across gateways and an edge coordinator.

A second line of work reduces calibration burden through transfer learning, domain alignment, and subject-adaptive decoding. Transfer learning is widely used in EEG-based BCI because EEG distributions vary across subjects, sessions, devices, and tasks \cite{wu2020transfer}. Alignment-based approaches make source and target EEG data more comparable before feature extraction or model adaptation, while MI-specific transfer pipelines combine alignment, spatial filtering, feature learning, and classifier adaptation to reduce the amount of new-user calibration required \cite{he2019transfer,wu2022transfer}. Recent reassessment of Euclidean alignment further reinforces the importance of distribution alignment in cross-subject and cross-session EEG transfer \cite{wu2025revisiting}. These methods motivate the use of reusable representations and limited target-subject adaptation. Beyond alignment-based transfer, EEG representation pretraining can further reduce calibration burden by providing a stronger backbone before subject-specific adaptation. Prior work has shown that pretrained EEG representations can improve learning when labeled target data are limited and can transfer across subjects, datasets, and downstream tasks \cite{banville2021uncovering,kostas2021bendr}. More recent large-scale pretrained models, including LaBraM, EEGPT, and CBraMod, extend this principle through self-supervised objectives designed to learn transferable representations from heterogeneous EEG data \cite{jiang2024large,wang2024eegpt,wang2025cbramod}. MIRepNet complements this general-purpose direction with an MI-specific pretrained pipeline designed for rapid adaptation under small calibration budgets \cite{liu2026mirepnet}. Collectively, these developments support evaluating whether a backbone initialized from pooled Session-1 EEG can facilitate later head-only personalization with limited target-session data.

From a neural-engineering perspective, sustained MI-BCI use also requires reliable decoding across sessions with limited recalibration, particularly for users with low or variable control \cite{saeedi2016long,ahn2015performance,lee2019eeg}. Recent healthy-to-stroke transfer results further indicate that pretrained MI representations can support adaptation when clinical data are limited \cite{nagarajan2024transferring}. Wearable and edge-enabled BCI systems further motivate local, resource-aware operation beyond controlled laboratory settings \cite{arpaia2023paving,wang2020accurate,bian2024device}. Existing work, however, largely studies calibration reduction through transfer, alignment, representation initialization, or target-subject adaptation rather than how a shared EEG representation is maintained through intermittent gateway-to-edge synchronization and subsequently reused for local personalization.

\subsection{Federated and Personalized Learning for EEG and BCI}

Federated learning changes the collaboration model by allowing distributed users or sites to improve a model without centralizing raw EEG data. In EEG and BCI, federated transfer learning has shown that cross-subject information can be exploited in a distributed setting \cite{ju2020federated}. Recent MI-BCI federated learning further addresses non-IID client data through client-specific normalization and robust local optimization \cite{jia2024federated}. Other federated BCI work studies collaboration across heterogeneous EEG datasets and acquisition devices, showing that federation can improve performance when data sources differ in format, scale, and collection conditions \cite{liu2024aggregating}. More recent approaches extend this direction through structure-guided personalization for heterogeneous MI clients and cross-subject federation designed to improve generalization to unseen EEG users \cite{hang2025personalized,liu2025mixeeg}. Together, these works establish FL as a valuable learning paradigm for EEG and MI-BCI.

The limitation is that most EEG/BCI FL studies frame the problem primarily as distributed model training. The communication process is usually represented as a standard federation protocol rather than as a deployment variable that can alter personalized decoding. Questions such as which gateways participate, whether delayed updates should be buffered, when stale updates should be admitted, and whether backbone downloads should be avoided are largely outside the evaluation scope. This leaves an important gap for home MI-BCI, where gateway availability and synchronization behavior can determine which model version a subject uses and how fresh the shared representation is during personalization.

Personalized FL provides a useful modeling foundation for this gap. Methods based on personalization layers, meta-learned initialization, and shared representations with local heads show that client heterogeneity can be handled by separating global and local model components \cite{arivazhagan2019federated,fallah2020personalized,collins2021exploiting}. This idea aligns naturally with MI-BCI: a shared backbone can encode reusable EEG structure, while a subject-specific head can adapt the decision boundary from limited calibration data. Our work adopts this principle, but evaluates it in a gateway-coordinated setting where the shared backbone is not assumed to be synchronized ideally. The central issue is not only how to personalize, but whether personalization remains reliable when the shared representation is refreshed through heterogeneous and policy-controlled communication.

\subsection{Communication-Aware, Asynchronous, and Edge Federated Learning}

The FL systems literature has developed a rich set of mechanisms for reducing communication cost and handling heterogeneous clients. FedAvg established iterative model averaging as a practical foundation for decentralized model learning \cite{mcmahan2017communication}. FedProx addresses statistical and systems heterogeneity by modifying the local optimization objective \cite{li2020federated}. Adaptive FL controls the tradeoff between local computation and global aggregation under edge resource constraints \cite{wang2019adaptive}. Client-selection systems such as Oort prioritize participants based on utility and system capability to improve time-to-accuracy \cite{lai2021oort}. Asynchronous and buffered approaches, including FedAsync-style optimization and FedBuff, reduce the need to wait for fully synchronized rounds and provide mechanisms for handling delayed updates \cite{xie2019asynchronous,nguyen2022federated}. More recent systems combine compute-aware semi-asynchronous scheduling with on-demand model broadcasting to reduce straggler delays, update staleness, and communication under heterogeneous or personalized deployments \cite{li2024fedcompass,li2024echopfl}. Hierarchical FL further introduces intermediate edge aggregation to reduce cloud backhaul, latency, and end-device energy cost \cite{liu2020client,abdellatif2022communication}.

These systems contributions are closely related to the mechanics of gateway-based MI-BCI, but their objectives are different. They are usually evaluated through convergence behavior, time-to-accuracy, communication rounds, scalability, or aggregate model performance on general machine-learning workloads. MI-BCI deployment imposes a narrower and more demanding interpretation of communication efficiency. A policy that reduces traffic can still be unacceptable if it reduces the reliability of feedback for subjects with weak links. A policy that maximizes freshness can be impractical if it requires frequent server-to-client backbone transfers. A policy that admits delayed updates can improve participation while also increasing the risk of stale contributions. Therefore, communication mechanisms must be evaluated not only as FL systems optimizations, but as determinants of MI-BCI personalization quality.

\subsection{Positioning of This Work}

This paper connects the above threads in a deployment-oriented MI-BCI setting. MI decoding and transfer-learning studies provide strong backbones and motivate limited-calibration adaptation. EEG/BCI FL studies show that collaboration can improve decoding while keeping raw EEG local. Personalized FL motivates the shared-backbone and subject-specific-head structure. Communication-aware FL systems provide mechanisms for participation control, buffering, staleness handling, and edge coordination.

The distinction is that this paper evaluates these mechanisms through the requirements of gateway-coordinated MI-BCI personalization. Unlike model-centered MI-BCI FL work, it treats synchronization behavior as part of the learning system rather than a fixed background assumption. Unlike generic communication-efficient FL work, it evaluates communication policy using MI-BCI outcomes: session-based personalization accuracy, client-to-server update cost, server-to-client backbone synchronization, update staleness, dropped updates, avoided downloads, link-availability sensitivity, and subject-level reliability. This positioning makes the work a deployment-centered study of federated MI-BCI personalization, in which learning performance and communication behavior are evaluated jointly.

In NEXUS-MI, data locality applies to the gateway-coordinated personalization and synchronization phase; EIB-PH uses pooled Session-1 data for predeployment backbone initialization, while formal privacy mechanisms such as secure aggregation, differential privacy, and encrypted computation are not evaluated.


\section{System Model and Problem Formulation}
\label{sec:system_model}

\subsection{Gateway-Coordinated MI-BCI Setting}

We consider an edge-enabled motor-imagery brain-computer interface (MI-BCI) system with a set of subjects $\mathcal{S} = \{1,\ldots,N\}$, where $N$ is the number of subjects. Each subject $s\in\mathcal{S}$ is served by a local gateway connected to a wearable EEG acquisition device. The gateway performs local inference for latency-sensitive decoding, stores the subject's EEG trials and calibration labels, and carries out subject-specific adaptation. During the gateway-coordinated personalization and synchronization phase, raw EEG trials, calibration labels, and subject-specific classifier heads remain at the gateway. Each gateway uploads a locally trained backbone parameter delta together with the metadata required for version-aware admission and reconstruction, while the coordinator may return a refreshed backbone state. The edge coordinator maintains the cohort-level shared backbone and coordinates model synchronization across gateways.

Because this study uses a one-subject/one-gateway abstraction, subject and gateway indices are in one-to-one correspondence. We use $s$ for subject data and decoder quantities, and $i$ only when emphasizing the communication state. Thus, subject-indexed quantities such as $\mathcal{D}^{(1)}_s$, $\mathcal{D}^{(2)}_s$, and $\phi_s$ describe the user's data and personalized classifier head, whereas gateway-indexed quantities such as $a_i^r$, $v_i^r$, and communication buffers describe synchronization state.

The gateway is the operational unit of the deployment. For each subject $s$, all subject-specific data and decoder components remain at the subject-side gateway. The edge coordinator does not receive raw EEG trials, calibration labels, or personalized classifier heads. Instead, it receives only backbone-related information produced by local training. This design allows each gateway to adapt to its subject while still allowing the shared representation to benefit from multi-subject learning. The resulting system is neither purely local nor fully centralized: gateways retain subject-specific data and personalization, while the edge coordinator maintains the shared representation used across the cohort.

In the present evaluation, this architecture is instantiated as a deployment model over public session-based MI-EEG datasets rather than as a real-time wearable implementation. Each subject is treated as a gateway-associated user, and recorded sessions are replayed to study how synchronization policy affects later-session personalization.

\subsection{Session-Based Personalization Protocol}

The system follows a session-based protocol that reflects repeated MI-BCI use. For each subject $s$, the available data are divided into an earlier session and a later session. The earlier session provides data for learning a shared representation and an initial subject-specific decoder state. The later session emulates continued use, where only a limited number of labeled calibration trials are available before evaluating the personalized decoder on held-out trials.

Let $\mathcal{D}^{(1)}s$ denote the earlier-session data for subject $s$, and let $\mathcal{D}^{(2)}s$ denote the later-session data. From $\mathcal{D}^{(2)}s$, the gateway forms a limited calibration set $\mathcal{C}{s,k}$ with calibration budget $k$, and a disjoint held-out evaluation set $\mathcal{T}{s,k}$. The calibration set is used for subject-local personalization at the gateway. No held-out trials from $\mathcal{T}{s,k}$ are used during gateway-side personalization. The held-out set is used only to measure later-session decoding after personalization.

This protocol separates two learning roles. The earlier session supports cohort-level representation learning across subjects. The later session tests whether that representation remains useful when each gateway performs limited-calibration personalization for its own subject. The protocol, therefore, evaluates continued-use personalization rather than only training-time classification performance.

\subsection{Backbone--Head Model Decomposition}

Each subject-specific decoder is decomposed into a shared EEG feature backbone and a personalized classifier head. Let $g_{\theta}$ denote the backbone with parameters $\theta$, and let $h_{\phi_s}$ denote the classifier head for subject $s$, with parameters $\phi_s$. The local decoder at gateway $s$ is
\begin{equation}
f_s(x)=h_{\phi_s}\bigl(g_{\theta_s}(x)\bigr),
\label{eq:local_decoder}
\end{equation}
where $\theta_s$ denotes the backbone state currently available at the gateway serving subject $s$. The backbone captures EEG representations that can benefit from multi-subject learning, whereas the classifier head represents the subject-specific decision boundary.

In the evaluated EEGNet configuration, $g_{\theta}$ comprises the two feature-extraction blocks preceding the terminal classifier. The first block applies eight temporal filters with a $1\times125$ kernel, batch normalization, and a depthwise spatial convolution with depth multiplier two spanning all EEG channels, followed by batch normalization, exponential linear activation, $1\times4$ average pooling, and dropout. The second block applies a $1\times22$ depthwise temporal convolution and a $1\times1$ pointwise convolution producing 16 feature maps, followed by batch normalization, exponential linear activation, $1\times8$ average pooling, and dropout. For the 1,000-sample input windows used in both datasets, $h_{\phi_s}$ consists of the terminal convolution that maps the 16 backbone feature maps to the dataset-specific classes using a $1\times31$ kernel, followed by log-softmax normalization. This decomposition yields 2,040 trainable backbone parameters and 1,988 trainable head parameters for BCICIV-2a, and 2,008 backbone parameters and 994 head parameters for OpenBMI, corresponding to 4,028 and 3,002 trainable parameters in total, respectively.

The edge coordinator maintains the shared backbone state, while each gateway maintains a local backbone copy and a subject-specific head. At each Session-1 local update, the gateway combines its currently available backbone with the common initial head state and jointly optimizes both components using the subject's Session-1 data. Joint optimization allows the subject-specific classification objective to shape the resulting backbone update. After local training, only the backbone-related update and its version metadata are eligible for synchronization with the edge coordinator; the trained classifier head remains at the gateway and replaces the previously retained head. 
Successive Session-1 local updates reuse the common head initialization. This limits carryover of head–backbone co-adaptation across communication rounds, so each backbone update is learned from the gateway’s current backbone state under a consistent classifier initialization rather than from a classifier shaped by earlier backbone states and participation history. The most recently trained head is retained locally to initialize Session-2 personalization.

During Session-2 personalization, each gateway combines the final collaborative backbone with the subject's most recently retained Session-1 head and adapts that head using $\mathcal{C}_{s,k}$. The backbone's trainable parameters are held fixed during this stage, and only the classifier-head parameters are optimized. Personalization therefore begins from a subject-specific head learned during Session 1 rather than from a randomly initialized classifier.

This decomposition is central to the system model because it separates cohort-level representation learning from subject-local adaptation. A gateway communication policy can alter the backbone version available locally, the freshness of updates reaching the coordinator, and the frequency with which refreshed backbone states return to the gateway. These effects can influence subsequent personalization even when the local calibration budget and the underlying EEG data remain fixed.

\subsection{Gateway-to-Edge Communication Model}

Communication occurs over synchronization rounds. At round $r$, the edge coordinator maintains a global backbone version $V^r$ with parameters $\theta^r$. Gateway $i$ maintains a local backbone version $v_i^r$ with parameters $\theta_i^r$. The difference $V^r-v_i^r$ represents the gateway's version lag relative to the coordinator.

Gateway availability is heterogeneous and time-varying. Let $a_i^r\in{0,1}$ denote the gateway-to-edge link state for gateway $i$ at round $r$. If $a_i^r=1$, the gateway is available for synchronization with the edge coordinator. If $a_i^r=0$, the gateway is unavailable for edge communication in that round. An unavailable gateway may continue local operation using its cached model state, but it cannot upload an update or download a refreshed backbone until communication becomes available.

The communication loop has two directions. In the client-to-server direction, a gateway uploads a backbone update expressed relative to the backbone version from which local training began, together with metadata identifying that base version. These uploads determine which subjects contribute to the next shared backbone. In the server-to-client direction, the coordinator sends a refreshed backbone state to a gateway. These downloads determine which backbone version is available for subsequent local training and personalization. Subject-specific classifier heads are not exchanged in this loop; they remain local to the gateways.

Intermittent communication can desynchronize the backbone versions held by the coordinator and the gateways. An update computed at a gateway may reach the coordinator after the global backbone has advanced, and a gateway may continue local operation using a cached backbone until a newer shared state is received. These conditions define the version-lag and update-freshness variables used by the coordination policies in Section~\ref{sec:methodology}; the specific rules for buffering, stale-update admission, and backbone downloads are introduced there.

We denote a gateway coordination policy by $\pi$. At each round, $\pi$ acts on the system variables defined above, including gateway availability, local and global backbone versions, pending updates, and gateway participation. The specific policy mechanisms are defined in Section~\ref{sec:methodology}; this section establishes the general variables and constraints on which those mechanisms operate.

\subsection{Problem Statement}

Consider a set of subject-side gateways with local MI-EEG sessions, limited later-session calibration data, and an edge coordinator that maintains a shared EEG backbone under heterogeneous gateway-to-edge connectivity. We study how gateway-coordination policy affects personalized MI-BCI performance. For a fixed dataset, learning regime, and calibration budget, each communication policy $\pi$ induces the operating point
\begin{equation}
\label{eq:policy_operating_point}
\begin{aligned}
\mathbf{z}(\pi)
=
\Bigl(
&\{A_s(\pi)\}_{s\in\mathcal{S}},
C^{\uparrow}(\pi),
C^{\downarrow}(\pi),
R_{\mathrm{rej}}(\pi), \\
&\overline{L}_{\mathrm{acc}}(\pi),
\overline{D}_{\mathrm{buf}}(\pi),
\{\Delta_s(\pi)\}_{s\in\mathcal{S}}
\Bigr),
\end{aligned}
\end{equation}
where $A_s(\pi)$ is the held-out later-session accuracy of subject $s$; $C^{\uparrow}(\pi)$ and $C^{\downarrow}(\pi)$ are the total client-to-server model-update traffic and server-to-client backbone traffic; $R_{\mathrm{rej}}(\pi)$ is the fraction of transmitted uploads rejected by the coordinator because their base checkpoint is unavailable or their backbone-version lag exceeds the stale-update threshold; $\overline{L}_{\mathrm{acc}}(\pi)$ is the event-weighted mean backbone-version lag of updates that pass coordinator admission and enter aggregation; and $\overline{D}_{\mathrm{buf}}(\pi)$ is the event-weighted mean communication-round delay of buffered uploads that pass the checkpoint-availability check and reach stale-update admission. Subject-level change relative to the ideal-link reference is defined as
\begin{equation}
\label{eq:subject_reliability_change}
\Delta_s(\pi)
=
A_s(\pi)-A_s^{\mathrm{ideal}}.
\end{equation}

This formulation treats gateway coordination as part of the federated MI-BCI learning system rather than as an independent communication overhead. The policies are compared as alternative operating points rather than optimized through a single scalar objective, because communication savings, coordinator rejection, accepted-update staleness, and buffered-upload delay may have different accuracy effects across datasets, learning regimes, and subjects. Subject-level reliability is assessed using both absolute personalized accuracy and change relative to the ideal-link reference, since aggregate performance can conceal substantial degradation or persistently low decoding accuracy for individual users.

\section{Gateway-Coordinated Federated Personalization}
\label{sec:methodology}

\textit{NEXUS-MI} operationalizes gateway-coordinated MI-BCI personalization through two design choices: a backbone--head learning structure and a communication policy that governs synchronization between gateways and the edge coordinator. The learning structure determines how cohort-level representation learning and subject-local adaptation are separated. The communication policy determines how backbone updates are delivered, admitted, and redistributed when gateway availability is heterogeneous. Section~IV-A defines the federated personalization regimes; Section~IV-B defines the synchronization conditions; Section~IV-C summarizes the policy taxonomy; Section~IV-D specifies the six gateway coordination policies; Section~IV-E details communication-aware gateway scheduling; Section~IV-F describes buffering, checkpoint retention, and stale-update admission; and Section~IV-G defines stale-aware backbone downloads. The framework builds on established federated-learning mechanisms for gateway participation, buffering, stale-update handling, and version-aware model delivery. Its contribution is to organize these mechanisms as a coordinated policy space for MI-BCI personalization and to evaluate how their combinations shape later-session decoding accuracy, communication cost, update freshness, and subject-level reliability under heterogeneous gateway-to-edge connectivity, rather than treating synchronization as a fixed communication assumption.

\begin{figure*}[h]
    \centering  \includegraphics[width=1.0\textwidth]{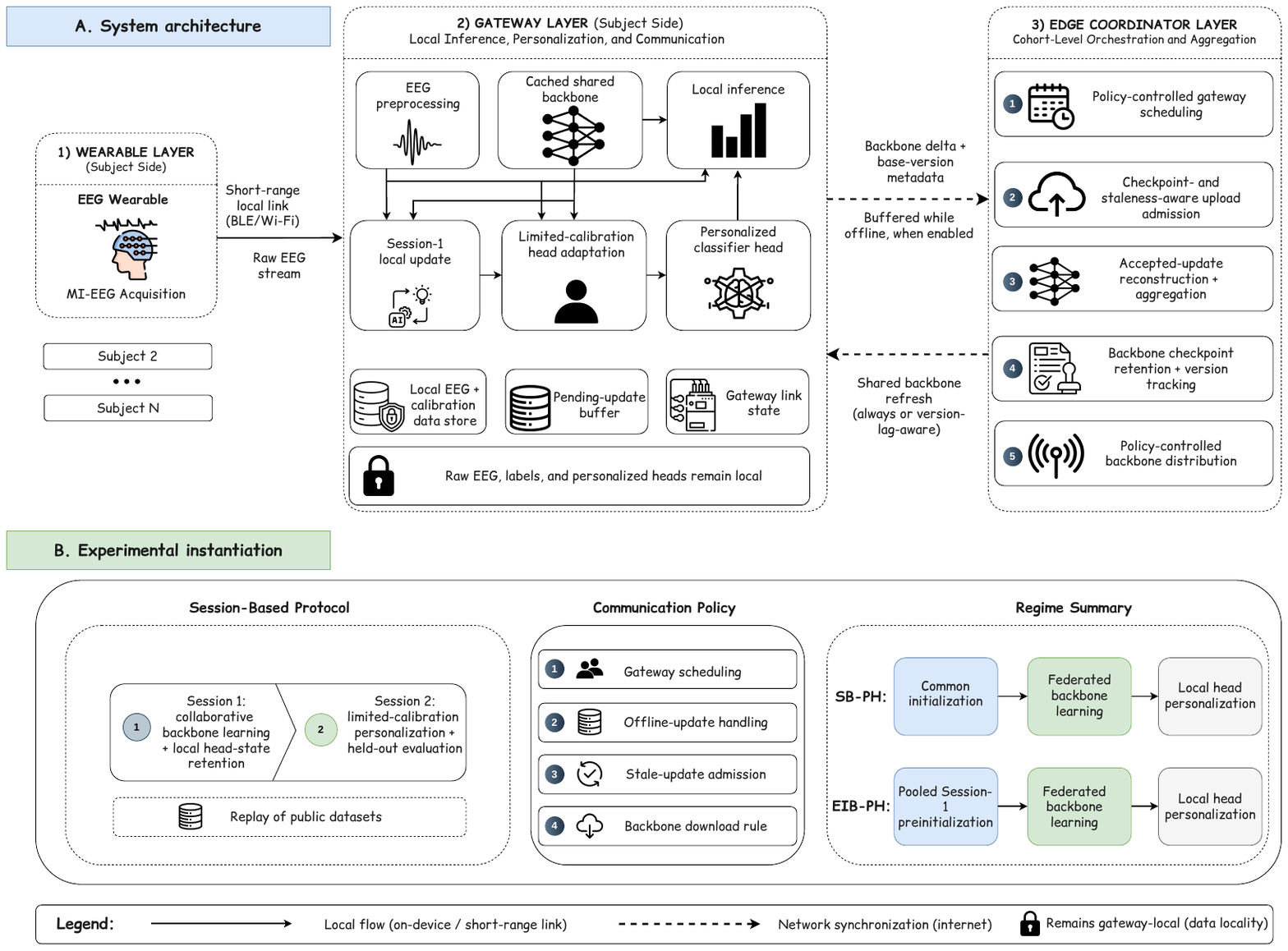}
    \caption{System architecture and experimental instantiation of NEXUS-MI.
\textbf{(A)} The system follows a three-tier wearable--gateway--edge design for communication-aware federated MI-BCI personalization. The wearable acquires MI-EEG signals and streams them to a subject-side gateway, where EEG preprocessing, local inference, Session-1 local updates, limited-calibration head adaptation, local data storage, link monitoring, and update buffering are performed. Raw EEG, calibration labels, and personalized classifier heads remain local to the gateway. The gateway communicates backbone deltas and associated base-version metadata to the edge coordinator, which performs policy-controlled gateway scheduling, checkpoint- and staleness-aware upload admission, accepted-update reconstruction and aggregation, backbone checkpoint retention and version tracking, and policy-controlled backbone distribution according to the active download rule.
\textbf{(B)} The architecture in \textbf{$(A)$} represents the intended deployment workflow, whereas the experiments instantiate this workflow through replay of public session-based MI-EEG datasets rather than a live online deployment. Session 1 supports collaborative backbone learning, while Session 2 uses limited calibration for local head personalization followed by held-out evaluation. The communication policy is defined by four synchronization dimensions: gateway scheduling, offline-update handling, stale-update admission, and backbone download control; the detailed policy configurations are provided in Table~I. SB-PH follows common initialization, federated backbone learning, and local head personalization, whereas EIB-PH uses pooled Session-1 preinitialization followed by the same federated backbone-learning and local head-personalization pipeline.}
    \label{fig4}
\end{figure*}

\subsection{Federated Personalization Regimes}
We evaluate two collaborative personalization regimes based on the decoder decomposition in Eq.~\ref{eq:local_decoder}. In both regimes, the backbone provides the shared EEG representation, while the classifier head remains subject-specific and is personalized at the gateway during later-session adaptation.

\textbf{Shared Backbone with Personalized Head (SB-PH).}
SB-PH learns a shared backbone through collaborative training across subject gateways. After collaborative backbone learning, each gateway personalizes only its classifier head. This regime evaluates whether a cohort-learned representation can support limited-calibration subject adaptation while retaining a local decision boundary.

\textbf{Embedding-Initialized Backbone with Personalized Head (EIB-PH).}
EIB-PH evaluates whether gateway-coordinated personalization benefits from a stronger initial EEG representation. In the present evaluation, the backbone is first initialized in a pre-deployment representation-learning step using pooled Session~1 data from the training subjects. The learned backbone weights are then used to initialize the subsequent gateway-coordinated training, communication-policy evaluation, and later-session head-only personalization procedure. Thus, EIB-PH changes the starting representation, but not the later-session calibration budget, personalization rule, communication policy, or evaluation protocol.

After this initialization step, EIB-PH follows the same gateway-local structure as SB-PH. Raw EEG trials, calibration labels, and subject-specific classifier heads remain at the subject-side gateways during the gateway-coordinated phase; the edge coordinator exchanges only backbone-related updates, version metadata, and policy-level synchronization state. The pooled initialization, therefore, defines the starting representation for EIB-PH, while the subsequent experiments evaluate how gateway synchronization policies affect personalization from that common initialization.

This regime separates two methodological questions. SB-PH measures how well a shared backbone can be learned directly through the federated personalization pipeline. EIB-PH measures whether initializing that backbone with a stronger cohort-level representation improves subsequent personalization and changes sensitivity to gateway synchronization. Reporting SB-PH and EIB-PH separately prevents conflating the effect of representation initialization with that of gateway coordination.

For both regimes, later-session personalization starts from the final collaborative backbone. During Session~2 personalization, the gateway keeps this backbone fixed and updates only the subject-specific classifier head using the limited calibration set $\mathcal{C}_{s,k}$. For each subject, the classifier head is initialized from the latest locally stored Session~1 head produced by that subject during collaborative training. If no Session~1 head state is available for a subject, personalization falls back to the common initial head state. This models continued decoder use, in which later-session adaptation begins from an existing subject-specific state produced during Session~1 training whenever such a state is available.

\subsection{Synchronization Conditions}
\label{subsec:synchronization_conditions}

The gateway-coordinated personalization framework is evaluated under an ideal-link reference and under heterogeneous gateway-to-edge connectivity. The ideal-link reference uses the same collaborative personalization pipeline but removes communication-induced constraints: selected gateways are available, synchronize with the edge coordinator when required, train locally, and upload current-round updates without missed uploads, delayed delivery, stale-update rejection, or base-checkpoint unavailability caused by link loss.

Under heterogeneous connectivity, each gateway $i$ is assigned once to an availability group $\gamma_i \in \{\mathrm{high},\mathrm{moderate},\mathrm{low}\}$ with online probability $p_{\gamma_i}$. At each communication round $r$, the gateway's online/offline state is sampled independently across gateways and rounds as
\begin{equation}
a_i^r \sim \mathrm{Bernoulli}(p_{\gamma_i}),
\label{eq:gateway_availability}
\end{equation}
where $a_i^r=1$ denotes that gateway $i$ is online and $a_i^r=0$ denotes that it is offline. For matched policy comparisons, the same generated availability trace is used across policies, so each policy is evaluated under the same realized sequence of online and offline gateway states. The sampled state affects gateway-edge synchronization only. An offline gateway retains its cached model state and cannot upload an update or download a refreshed backbone until a later online round. When the active policy enables offline buffering, the gateway may also retain a locally produced pending update for later synchronization.

The main heterogeneous-link policy study uses the default availability setting: high-, moderate-, and low-availability gateways have online probabilities $0.95$, $0.70$, and $0.40$, respectively, with target group fractions $0.34$, $0.33$, and $0.33$. Because each dataset contains a finite number of subject gateways, these fractions define the intended group proportions, while the realized group counts are determined by the available subject count. These operating points are intended to emulate reliable home or institutional Wi-Fi, intermittent consumer connectivity, and weak or mobile backhaul. This abstraction follows prior cross-device FL work on availability, participant selection, and delayed aggregation~\cite{bonawitz2019towards,lai2021oort,nguyen2022federated}.

To avoid tying the conclusions to a single availability setting, the sensitivity study repeats the same high/moderate/low grouping under three severity levels. The mild setting uses online probabilities $0.98/0.85/0.60$, the default setting uses $0.95/0.70/0.40$, and the severe setting uses $0.90/0.50/0.20$ for the high-, moderate-, and low-availability groups, respectively.

\subsection{Policy Taxonomy}
\label{subsec}

All six heterogeneous-link settings are gateway coordination policies. They share the same link model but differ in four synchronization dimensions: scheduling, offline update rule, stale update rule, and download rule. Table~\ref{tab:communication_policies} gives the compact definitions used throughout the paper.

\begin{table*}[t]
\centering
\caption{Gateway coordination policies under heterogeneous gateway-to-edge links.}
\label{tab:communication_policies}
\resizebox{\textwidth}{!}{%
\begin{tabular}{p{0.06\textwidth} p{0.18\textwidth} p{0.20\textwidth} p{0.22\textwidth} p{0.20\textwidth}}
\toprule
\textbf{Policy} & \textbf{Scheduling} & \textbf{Offline update rule} & \textbf{Stale update rule} & \textbf{Download rule} \tabularnewline
\midrule
P1 & NA-all & Discard missed update & No delayed update & Download when selected \tabularnewline
P2 & NA-all & FIFO buffer & Accept if base retained & Download when selected \tabularnewline
P3 & NA-all & FIFO buffer & Drop if $\ell_u^r>\tau_s$ & Download when selected \tabularnewline
P4 & NA-all & Latest pending update & Drop if $\ell_u^r>\tau_s$ & Download when selected \tabularnewline
P5 & CA-priority & FIFO buffer & Drop if $\ell_u^r>\tau_s$ & Refresh if $V^r-v_i^r>\tau_d$ \tabularnewline
P6 & CA-priority & Latest pending update & Drop if $\ell_u^r>\tau_s$ & Refresh if $V^r-v_i^r>\tau_d$ \tabularnewline
\bottomrule
\end{tabular}%
}

\vspace{0.4em}
\begin{minipage}{0.98\textwidth}
\footnotesize
\emph{Notes:} NA-all denotes non-adaptive all-gateway scheduling, in which gateways are selected before online/offline availability is realized. CA-priority denotes communication-aware scheduling over currently online gateways using the priority rule in Eq.~\ref{eq:priority_tuple}. FIFO buffers are bounded with $B_{\max}=3$ in the experiments. The latest pending update retains only the most recent delayed update for a gateway. The stale-update threshold is $\tau_s=2$ and is applied to the delayed-update lag $\ell_u^r$ defined in Eq.~\ref{eq:update_staleness}. The download threshold is $\tau_d=1$ and is applied to the gateway's local backbone lag. P3 and P5 share the same FIFO offline-update rule and stale-update rule; they differ only in the scheduling and download-rule columns.
\end{minipage}
\end{table*}


The \textbf{scheduling} dimension determines which gateways receive immediate upload opportunities. Under non-adaptive all-gateway scheduling, every gateway is scheduled in each communication round before the realized link state is applied. Under communication-aware gateway scheduling, the coordinator first observes the realized online set and assigns immediate upload opportunities only to gateways that can communicate in that round.

The \textbf{offline update rule} determines what happens when a gateway produces an update but cannot immediately upload it. The update may be discarded, stored in a bounded FIFO buffer, or used to replace the gateway's previous pending update under the latest-update rule. The \textbf{stale update rule} determines whether a delayed update is admitted after it later reaches the coordinator. The update may be rejected because no delayed update was retained, admitted if its base checkpoint is still available, or rejected if its version lag exceeds the stale threshold $\tau_s$. The \textbf{download rule} determines whether an online gateway receives a refreshed backbone from the coordinator. Always-download synchronization sends the newer backbone whenever one is available, while stale-aware synchronization sends it only when the gateway's cached backbone is sufficiently behind the coordinator.

\subsection{Gateway Coordination Policies}
\label{sec:gateway-policies}

The policies form two families. P1--P4 provide the non-adaptive reference family under heterogeneous gateway availability. In these policies, all gateways are scheduled before the online/offline state is applied; only gateways that are online in the realized link state can synchronize in that round. Differences within this family therefore isolate how missed updates, buffering, and stale-update admission affect collaborative backbone learning when scheduling itself does not adapt to availability.

P5--P6 provide the communication-aware family. These policies first observe the currently online gateway set and assign immediate upload opportunities only to gateways that can communicate in the current round. They also use stale-aware backbone downloading, so a gateway refreshes its cached backbone only when its local version lag exceeds the download threshold. The difference between P5 and P6 is the offline-update rule: P5 retains delayed updates in a bounded FIFO buffer, whereas P6 retains only the latest pending update from each offline gateway.

This policy organization makes P3 and P5 the principal controlled comparison. Both policies use FIFO buffering and stale-update rejection, so delayed-update handling is held fixed. They differ in synchronization behavior: P3 uses non-adaptive all-gateway scheduling with always-download synchronization, whereas P5 combines communication-aware gateway scheduling with stale-aware backbone downloading. The P3/P5 comparison, therefore, isolates the effect of communication-aware synchronization that adapts gateway selection and backbone refresh to realized availability and version lag while holding the buffering and stale-update admission rules constant.

\subsection{Communication-Aware Gateway Scheduling}
\label{subsec:communication_aware_scheduling}

P5 and P6 use an online-first scheduling rule. At the beginning of communication round $r$, the heterogeneous link model determines the online and offline gateway sets:
\begin{equation}
\label{eq:online_offline_sets}
\mathcal{O}_r=\{i : a_i^r=1\}, \qquad
\mathcal{F}_r=\{i : a_i^r=0\}.
\end{equation}
Only gateways in $\mathcal{O}_r$ are eligible for immediate upload. Let $K$ denote the per-round gateway-selection budget; in the experimental protocol, $K=6$ for BCICIV-2a and $K=40$ for OpenBMI, as summarized in Table~\ref{tab:experimental-protocol}. If $|\mathcal{O}_r|\leq K$, all online gateways are selected. If $|\mathcal{O}_r|>K$, the coordinator ranks online gateways by synchronization need and selects the highest-priority gateways. The priority ordering is a fixed scheduling heuristic rather than the solution of an explicit optimization problem; its intuition is to favor gateways that have not uploaded successfully recently, carry buffered updates at greater risk of becoming stale, or hold a more outdated local backbone.

For each online gateway $i$, the coordinator computes the priority vector
\begin{equation}
\label{eq:priority_tuple}
q_i(r)=
\begin{pmatrix}
\Delta^{\mathrm{succ}}_i(r)\\
T^{\mathrm{succ}}_i(r)\\
B_i(r)\\
S^{\mathrm{buf}}_i(r)\\
D^{\mathrm{buf}}_i(r)\\
\Delta^{\mathrm{loc}}_i(r)
\end{pmatrix}.
\end{equation}
Here, $\Delta^{\mathrm{succ}}_i(r)$ is the coordinator-version lag since gateway $i$'s last successful upload, $T^{\mathrm{succ}}_i(r)$ is the number of rounds since that upload, $B_i(r)$ indicates whether the gateway has a nonempty pending-update buffer, $S^{\mathrm{buf}}_i(r)$ and $D^{\mathrm{buf}}_i(r)$ are the maximum staleness and delay among buffered payloads, and $\Delta^{\mathrm{loc}}_i(r)$ is the lag between the current coordinator backbone and the gateway's cached local backbone.

The coordinator sorts online gateways in descending lexicographic order of $q_i(r)$. At the start of training, gateways have no prior successful upload, so the last successful upload round and version are initialized to zero, and the corresponding lag terms are measured from the beginning of the collaborative process. For a gateway with no pending update, $B_i(r)=0$, $S_i^{\mathrm{buf}}(r)=-1$, and $D_i^{\mathrm{buf}}(r)=0$. Ties are resolved using a seeded round-specific random key, followed by the subject identifier, so matched policy comparisons remain reproducible. Because the ordering is lexicographic, earlier components intentionally dominate later components: version lag since the last successful upload is prioritized before time since the last successful upload, pending-buffer status, buffered-update staleness, buffered-update delay, and local-backbone lag. The number of selected gateways is
\begin{equation}
\label{eq:selected_gateway_count}
m_r=\min\{K,|\mathcal{O}_r|\},
\end{equation}
and the selected set contains the top $m_r$ online gateways under the priority ordering.

The scheduling rule defines which gateways perform an immediate upload in a communication round. Under non-adaptive all-gateway policies, every gateway is scheduled before availability is applied. A scheduled online gateway may refresh its cached backbone according to the active download rule, train locally, and upload a fresh update. A scheduled offline gateway trains from its cached backbone, and the resulting update is discarded or retained according to the active offline-update rule. Under communication-aware policies, only selected online gateways receive immediate upload opportunities. A selected online gateway first delivers any pending buffered updates and then trains and uploads a fresh update from its current cached or refreshed backbone. Each buffered payload and the subsequently generated fresh payload are transmitted as separate update instances and are evaluated independently by the coordinator. Consequently, a gateway may contribute multiple accepted update instances in the same communication round. Offline gateways are not selected for immediate upload; under P5 and P6, they may continue local training from their cached backbones and retain the resulting updates according to the active buffering rule. Online gateways that are available but not selected within the gateway-selection budget remain idle and do not train, upload, or download in that communication round. The local training objective, update reconstruction, aggregation rule, buffering behavior, stale-update admission, and backbone-download rule remain governed by the active communication policy.

\subsection{Buffering, Checkpoints, and Stale-Update Admission}
\label{subsec:buffering_stale_admission}

When a gateway cannot deliver an update in the current communication round, the active offline update rule determines whether the locally produced update is retained for later synchronization. With no buffering, the missed update is discarded. With FIFO buffering, pending updates are stored up to a bounded capacity $B_{\max}$, and the oldest pending update is dropped when the buffer overflows. With latest-update buffering, only the most recent pending update is retained for that gateway, replacing any older pending update. Bounded buffering is instantiated with capacity $B_{\max}=3$, allowing each gateway to recover from short offline periods while preventing unbounded queues of delayed and increasingly stale updates.

A delayed update is interpretable only with respect to the backbone checkpoint from which it was trained. In the gateway-to-coordinator direction, each communicated learning payload consists of a backbone delta and its associated metadata, including the coordinator version from which the local training step began. The coordinator, therefore, maintains a versioned checkpoint history so that delayed backbone deltas can be reconstructed as candidate backbone states before aggregation. For policies that apply stale-drop admission, this history is finite. After the coordinator advances to version $V^r$, it retains checkpoint versions
\begin{equation}
\label{eq:checkpoint_retention}
v \geq V^r-\tau_s-\rho .
\end{equation}
where $\tau_s$ denotes the stale-update admission threshold and $\rho$ denotes the checkpoint-retention margin. The margin controls reconstruction availability, while the stale-update threshold controls admission. This separation allows delayed updates to be reconstructed from their base checkpoints before the coordinator determines whether they are admissible under the staleness rule. We set $\rho=5$ for all experiments and keep it fixed across datasets, subjects, and stale-drop policies. This value provides a reconstruction margin larger than the stale-admission threshold, allowing recently delayed payloads to be reconstructed from their base checkpoints while keeping the retained checkpoint history bounded. With $\tau_s=2$, the retained history contains at most eight checkpoint versions. For accept-if-base admission, the checkpoint history is not pruned by the stale-drop horizon, since delayed updates are screened by base-checkpoint availability rather than by staleness.

When a delayed update later reaches the coordinator, the coordinator first checks whether the corresponding base checkpoint is still retained. If the base checkpoint is unavailable, the update is rejected as checkpoint-missing. Such drops occur when a delayed upload refers to a base version that has aged out of the retained checkpoint history. If the base checkpoint is available, the update's staleness is computed as
\begin{equation}
\label{eq:update_staleness}
\ell_u^r = V^r - v_u .
\end{equation}
Here, $V^r$ is the coordinator backbone version at round $r$, and $v_u$ is the backbone version from which the delayed update $u$ was trained. Accept-if-base admission admits delayed updates whose base checkpoint is available. Stale-drop admission further rejects delayed updates satisfying
\begin{equation}
\label{eq:stale_drop_rule}
\ell_u^r > \tau_s .
\end{equation}

Stale-drop admission is instantiated with a fixed threshold $\tau_s=2$ for all subjects, allowing modest delayed delivery while rejecting updates trained from backbone versions that are too far behind the coordinator. For an accepted update $u$, let $v_u$ denote the coordinator backbone version from which the gateway trained, and let $\Delta\theta_u$ denote the transmitted backbone delta. The corresponding candidate backbone state is reconstructed as
\begin{equation}
\label{eq:backbone_reconstruction}
\tilde{\theta}_u^r
=
\theta^{v_u}+\Delta\theta_u .
\end{equation}
This reconstruction makes each delayed update interpretable relative to the retained base checkpoint from which it was produced.

Accepted candidate backbone states are aggregated at the update-instance level using a uniform averaging rule that is held fixed across all policies. Because buffered and fresh payloads are evaluated independently, a gateway that contributes multiple accepted payloads in a communication round contributes multiple candidate backbone states to that round's aggregation. Gateway scheduling, buffering, and coordinator admission therefore determine the composition of the accepted-update set. The shared backbone is updated as
\begin{equation}
\label{eq:backbone_aggregation}
\theta^{r+1}
=
\frac{1}{|\mathcal{A}_r|}
\sum_{u\in\mathcal{A}_r}
\tilde{\theta}_u^r,
\qquad
|\mathcal{A}_r|>0.
\end{equation}
Here, $\mathcal{A}_r$ denotes the set of accepted update instances at round $r$, and each accepted buffered or fresh payload contributes one reconstructed candidate backbone state $\tilde{\theta}_u^r$. As defined in Eq.~\eqref{eq:backbone_reconstruction}, each candidate state is reconstructed from the retained checkpoint corresponding to the base version identified in the payload metadata. If $\mathcal{A}_r$ is empty, no aggregation is performed; the coordinator retains its current backbone state and version for the next communication round. The same uniform update-instance aggregation rule is used across all policies, while scheduling, buffering, checkpoint availability, and stale-update admission determine which update instances enter the accepted set. Only backbone updates enter the aggregation; subject-specific classifier heads remain gateway-local.

\subsection{Stale-Aware Backbone Downloads}
\label{subsec:stale_aware_download}

The download rule determines when an online gateway refreshes its cached backbone from the edge coordinator before local training. Under always-download synchronization, a selected gateway downloads the coordinator backbone whenever the coordinator version is newer than the gateway's cached version. Under stale-aware synchronization, the gateway downloads only when its local backbone lag exceeds a fixed download threshold:
\begin{equation}
\label{eq:download_rule}
V^r - v_i^r > \tau_d ,
\end{equation}
where $V^r$ is the coordinator backbone version at round $r$ and $v_i^r$ is the backbone version currently cached by gateway $i$.

Stale-aware downloading is instantiated with threshold $\tau_d=1$ for all subjects. Because stale-update admission allows delayed updates up to $\tau_s=2$, the download rule triggers refresh before local training proceeds from a backbone that is already at the stale-update boundary. A gateway may therefore train with a recently cached backbone, but it refreshes once its local copy is more than one coordinator version behind. This choice reduces unnecessary server-to-gateway backbone transfers while limiting the synchronization lag from which subsequent local updates are produced. Together with $B_{\max}=3$ and $\tau_s=2$, the download threshold is fixed across datasets and subjects so that policy comparisons reflect coordination behavior rather than threshold tuning. Each policy is therefore treated as a multi-objective operating point over decoding accuracy, client-to-server traffic, server-to-client traffic, update freshness, and subject-level reliability, rather than as the optimizer of a single scalar objective.

\section{Experimental Evaluation}
\label{sec:evaluation}

This section evaluates gateway-coordinated federated personalization for motor-imagery brain-computer interfaces (MI-BCIs) under heterogeneous gateway-to-edge connectivity. The experiments examine communication policy as an element of the learning process rather than as an external transmission cost. In this setting, gateway availability, delayed update delivery, stale-update admission, and backbone synchronization determine which subject updates shape the shared representation and which backbone version is available for subsequent personalization. The evaluation, therefore, reports decoding accuracy together with communication volume, update rejection, accepted-update staleness, backbone download behavior, link-availability sensitivity, and subject-level reliability.

The analysis proceeds from reference behavior to constrained deployment behavior. We first establish the ideal-link performance of the two personalization regimes, SB-PH and EIB-PH, to define the accuracy and synchronization reference when gateway communication is unconstrained. We then evaluate the full P1--P6 policy landscape under the default heterogeneous-link profile, showing how participation, buffering, stale-update handling, and download control create distinct accuracy--communication operating points. To separate the effect of communication-aware coordination from the broader policy landscape, we compare P5 with the matched non-adaptive FIFO stale-rejection policy P3 and further decompose the communication-aware policy through a component analysis. We then assess the robustness of the principal P3/P5 contrast across five matched training and gateway-availability realizations. We next vary link availability from mild to severe conditions to test whether the observed trade-offs persist under changing connectivity. Finally, we examine subject-level outcomes relative to the ideal-link reference, since aggregate accuracy alone can conceal large individual gains, losses, and low-accuracy cases. During the gateway-coordinated phase, raw EEG, calibration labels, and subject-specific classifier heads remain at the subject-side gateways, and collaboration occurs through shared-backbone updates; EIB-PH additionally uses pooled Session-1 data for predeployment backbone initialization. The evaluation structure follows the research questions: the ideal-link reference addresses \textbf{RQ1}; the P1--P6 policy landscape, controlled P3/P5 comparison, and communication-aware component analysis address \textbf{RQ2}; and the link-availability sensitivity and subject-level reliability analyses address \textbf{RQ3}.

\subsection{Experimental Setting}
\label{sec:experimental-setting}

Table~\ref{tab:experimental-protocol} summarizes the experimental protocol. We evaluate BCICIV-2a and OpenBMI, two public MI-BCI datasets with complementary cohort sizes and class structures. Both datasets comprise benchmark MI-EEG recordings collected under controlled laboratory conditions, rather than recordings from rehabilitation patients or home-use deployments. BCICIV-2a contains 9 subjects and four motor-imagery classes, whereas OpenBMI contains 54 subjects and two classes. Each subject is represented by one subject-side gateway. For both datasets, Session~1 supports collaborative backbone learning, and Session~2 supports gateway-side personalization and held-out evaluation.

EEG trials are represented as four-second epochs sampled at 250~Hz. BCICIV-2a uses 22 EEG channels, and OpenBMI uses a selected 20-channel motor-imagery montage. All experiments use EEGNet as the shared feature-learning backbone, with dataset-specific input-channel and output-class dimensions. The EEGNet configuration uses $F_1=8$, depth multiplier $D=2$, $F_2=16$, temporal kernel length $C_1=125$, and dropout probability 0.5. Collaborative training uses 120 communication rounds, 50 local epochs for each newly produced gateway update, Adam optimization with learning rate $10^{-3}$, batch size 16, and FedAvg-style uniform averaging of accepted reconstructed backbone states at the update-instance level. Subject-specific classifier heads are not aggregated.

For Session~2 personalization, each gateway adapts only the classifier head using $k\in\{15,20,30\}$ labeled calibration trials per class. The chronological Session~2 split keeps the held-out test set fixed across calibration budgets. BCICIV-2a uses the first 144 Session~2 trials as the calibration pool and the remaining 144 trials for testing; OpenBMI uses the first 80 Session~2 trials as the calibration pool and the remaining 120 trials for testing. The Session~2 head is initialized from the subject's latest learned Session~1 head when available and is trained with learning rate $5\times10^{-5}$, a maximum of 750 epochs, and early-stopping patience of 100.

The ideal-link reference removes communication-induced missed uploads and delayed delivery. The default heterogeneous-link profile uses high-, moderate-, and low-availability probabilities of $0.95/0.70/0.40$, with target group fractions of $0.34/0.33/0.33$. Communication-aware scheduling uses gateway-selection budgets of $K=6$ for BCICIV-2a and $K=40$ for OpenBMI. For policies that enable the corresponding mechanisms, the buffer capacity is fixed at $B_{\max}=3$, stale-update rejection uses $\tau_s=2$ backbone versions, and stale-aware backbone downloading uses $\tau_d=1$ backbone version. To support controlled policy comparisons, all policies within a given realization use identical data partitions, Session~2 calibration samples, model initialization, gateway-group assignment, and realized availability trace. The complete P1--P6 landscape, component analysis, availability-severity study, and subject-level association analysis are reported for the primary realization using seed 2026. The principal P3/P5 contrast is additionally evaluated across five matched realizations with different training initializations and gateway-availability traces to determine whether its accuracy and communication effects persist beyond the primary realization; the full robustness analysis is reported in Supplementary Section~S2.

\begin{table*}[t]
\centering
\caption{Experimental setting for gateway-coordinated federated MI-BCI personalization.}
\label{tab:experimental-protocol}
\begingroup
\footnotesize
\setlength{\tabcolsep}{5pt}
\renewcommand{\arraystretch}{1.08}

\begin{tabular}{@{}p{0.22\textwidth}p{0.75\textwidth}@{}}
\toprule

\multicolumn{2}{c}{\textbf{A. Data and learning protocol}} \tabularnewline
\midrule

\textbf{Evaluation component} & \textbf{Configuration used in the experiments} \tabularnewline
\midrule

Datasets and gateways &
BCICIV-2a: 9 subjects and four classes; OpenBMI: 54 subjects and two classes. Each subject is modeled as one subject-side gateway. \tabularnewline

EEG preprocessing &
Four-second epochs sampled at 250~Hz. BCICIV-2a uses 22 EEG channels; OpenBMI uses a selected 20-channel motor-imagery montage. \tabularnewline

EEGNet configuration &
Shared EEGNet backbone with dataset-specific input and output dimensions; $F_1=8$, $D=2$, $F_2=16$, $C_1=125$, and dropout probability 0.5. \tabularnewline

Session protocol &
Session~1 supports collaborative backbone learning. Session~2 supports gateway-side head personalization and held-out testing. \tabularnewline

Learning regimes &
SB-PH uses a commonly initialized backbone for federated learning with subject-specific heads. EIB-PH initializes the backbone through pooled Session~1 pretraining and then follows the same federated-training and head-only personalization protocol. \tabularnewline

EIB-PH initialization &
Pooled Session~1 trials from the participating training subjects with a seeded random 80/20 class-stratified trial-wise split; subjects may contribute trials to both subsets, and validation is used for early stopping and model selection rather than unseen-subject evaluation. Adam optimizer; learning rate $10^{-3}$; maximum 1500 epochs; early-stopping patience 200. \tabularnewline

Federated training &
120 communication rounds; 50 local epochs for each newly produced gateway update; Adam optimizer; learning rate $10^{-3}$; batch size 16. \tabularnewline

Aggregation rule &
FedAvg-style uniform averaging of accepted reconstructed backbone states at the update-instance level. Subject-specific classifier heads remain gateway-local. \tabularnewline

Calibration sampling &
$k\in\{15,20,30\}$ labeled Session~2 calibration trials per class, selected chronologically from a fixed calibration pool. The last selected trial from each class is reserved for early-stopping validation, and the remaining selected trials are used for head updates. The held-out test set remains fixed across calibration budgets. \tabularnewline

Session~2 split &
BCICIV-2a: first 144 trials for the calibration pool and remaining 144 trials for testing. OpenBMI: first 80 trials for the calibration pool and remaining 120 trials for testing. \tabularnewline

Head personalization &
The available backbone is held fixed, and only the subject-specific classifier head is updated. The head is initialized from the latest locally stored Session~1 head, with the common initial head used if no Session~1 head is available; learning rate $5\times10^{-5}$; maximum 750 epochs; early-stopping patience 100. \tabularnewline

\midrule
\multicolumn{2}{c}{\textbf{B. Communication and statistical protocol}} \tabularnewline
\midrule

\textbf{Evaluation component} & \textbf{Configuration used in the experiments} \tabularnewline
\midrule

Communication reference &
Ideal-link reference with uninterrupted gateway availability, no communication-induced missed or delayed uploads, and immediate backbone synchronization when a newer coordinator version is available. \tabularnewline

Default heterogeneous links &
High-, moderate-, and low-availability probabilities of $0.95/0.70/0.40$, with target gateway-group fractions of $0.34/0.33/0.33$. \tabularnewline

Communication parameters &
Gateway-selection budget $K=6$ for BCICIV-2a and $K=40$ for OpenBMI; FIFO buffer capacity $B_{\max}=3$ updates per gateway; stale-update admission threshold $\tau_s=2$ backbone versions; stale-aware download threshold $\tau_d=1$ backbone version. \tabularnewline

Seed and availability traces &
Primary analyses use seed 2026. For robustness, the P3/P5 comparison is repeated across five matched realizations with distinct training initializations and gateway-availability traces. \tabularnewline

Metrics and units &
Held-out subject-level accuracy and reliability relative to the ideal-link reference; client-to-server model-update and server-to-client backbone traffic in decimal megabytes ($1~\mathrm{MB}=10^6$ bytes); coordinator-rejected uploads as a percentage of transmitted uploads; event-weighted mean accepted-update staleness in backbone versions; event-weighted mean buffered-upload delay in communication rounds; and download avoidance as a percentage of synchronization opportunities. \tabularnewline

Statistical unit &
Paired tests use subjects as the statistical unit after first averaging each subject's accuracy across the three calibration budgets. \tabularnewline

\bottomrule
\end{tabular}

\endgroup
\end{table*}

\subsection{Metrics and Statistical Methodology}
\label{sec:metrics-statistics}

The primary learning outcome is subject-level decoding accuracy after gateway-side head-only personalization. Accuracy is reported separately by calibration budget where relevant and averaged across $k\in\{15,20,30\}$ for aggregate policy comparisons. Communication is measured as client-to-server model-update traffic and server-to-client backbone traffic in decimal MB. Synchronization behavior is characterized by the coordinator-rejected upload rate, mean accepted-update staleness, mean buffered-upload delay, and download avoidance. The rejection rate is computed over transmitted uploads; accepted-update staleness is averaged over admitted uploads and reported in backbone versions; buffered-upload delay is averaged over buffered uploads that reach stale-update admission and reported in communication rounds; and download avoidance is measured over synchronization opportunities. Subject-level reliability is evaluated using absolute personalized accuracy and accuracy differences relative to the matched ideal-link reference, including median and extreme differences, low-accuracy counts, changes of at least five percentage points, and the number of subjects who are both below the task-specific analytical threshold and degraded by at least 5~pp relative to the matched ideal-link reference.

Statistical inference uses subjects as the unit of analysis. For each subject, policy, and learning regime, accuracy is first averaged across the three calibration budgets so that repeated measurements from the same subject are not treated as independent observations. Controlled policy comparisons are then based on paired subject-level differences. We report the mean paired accuracy difference in percentage points, its two-sided 95\% Student-$t$ confidence interval, the standardized paired effect size $d_z$, the two-sided paired $t$-test $p$-value, and the Wilcoxon signed-rank $p$-value. The paired effect size is
\[
d_z=\frac{\bar{\Delta}}{s_{\Delta}},
\]
where $\bar{\Delta}$ and $s_{\Delta}$ are the mean and standard deviation of the paired subject-level differences, respectively.

For the principal P5-versus-P3 comparison, Holm adjustment is applied across the four paired $t$-tests defined by the two datasets and two learning regimes. The link-availability analysis is treated as a sensitivity analysis: inference for the default profile is taken from the prespecified primary comparison, while the additional profile-specific comparisons are summarized using paired differences and 95\% confidence intervals. The Wilcoxon signed-rank tests are reported as complementary unadjusted analyses. Given the smaller BCICIV-2a cohort ($n=9$), its results are interpreted using confidence intervals and effect sizes alongside the paired tests. The complete P1--P6 operating points across accuracy, traffic, coordinator rejection, accepted-update staleness, and buffered-upload delay are reported in Supplementary \textbf{\textit{Table~S1}}.

Robustness of the principal P5-versus-P3 comparison is assessed across five matched realizations. The primary repeated-run interval is a crossed hierarchical bootstrap with 10{,}000 resamples of matched realizations and subjects while preserving the within-subject policy pairing. Across-realization $t$ intervals and observed ranges are reported descriptively, and each matched realization serves as the replicate unit for communication outcomes. Because several stochastic elements of the experimental realization vary across the matched replicates, the analysis characterizes their combined influence rather than isolating individual sources of run-to-run variability.

\subsection{Ideal-Link Reference Performance}
\label{sec:ideal-link}

The ideal-link reference establishes the behavior of the learning pipeline under unconstrained gateway-to-edge synchronization. In this condition, selected gateways synchronize when required, uploads are delivered without communication-induced delay, and refreshed backbone states are available without missed downloads. The purpose of this condition is to provide a stable synchronization reference for \textbf{RQ1} and a subject-specific baseline for the later reliability analysis. It is best interpreted as a reference synchronization condition rather than an accuracy ceiling. Ideal synchronization maximizes the regularity of model exchange, but it does not necessarily produce the highest empirical accuracy for every dataset, regime, or calibration budget.

This distinction is important because heterogeneous-link policies also change the learning trajectory. When synchronization is constrained, the coordinator may aggregate a different sequence of subject updates, admit delayed updates with bounded staleness, reject older updates, or refresh gateway backbones less frequently. These mechanisms can alter the effective optimization path of the shared backbone. In limited-calibration and subject-heterogeneous EEG learning, such changes can sometimes act as implicit regularization: fewer backbone refreshes may reduce sensitivity to round-to-round update variation, bounded stale updates may smooth or redirect the aggregation trajectory, and stochastic gateway availability may produce small empirical gains for some dataset-regime combinations. Therefore, heterogeneous-link accuracy can occasionally exceed the ideal-link reference, even though the ideal-link condition remains the cleanest reference for interpreting communication-constrained operation.
Supplementary Figure~S1 reports mean decoding accuracy across the three Session~2 calibration budgets for the ideal-link reference and Policies~P1--P6 under both learning regimes.

OpenBMI shows a clear separation between the two learning regimes under the ideal-link reference. SB-PH reaches 72.23\% mean accuracy, whereas EIB-PH reaches 80.75\%. This 8.52 percentage-point difference indicates that the pooled Session~1 representation initialization provides a substantial reference-setting advantage on this two-class cohort. On BCICIV-2a, the two regimes are closer: SB-PH reaches 67.77\%, while EIB-PH reaches 66.82\%. Thus, the value of the representation-initialized regime is dataset-dependent, with a clear ideal-link gain on OpenBMI and no corresponding ideal-link advantage on BCICIV-2a.

The ideal-link rows also establish the communication exposure of unconstrained collaboration. BCICIV-2a requires 9.30~MB of client-to-server traffic and 9.23~MB of server-to-client traffic. OpenBMI requires 55.00~MB and 54.56~MB, respectively. These values define the synchronization-cost reference against which heterogeneous-link policies are interpreted. Accordingly, the heterogeneous-link policies are interpreted as joint learning-and-synchronization operating points that couple personalized decoding accuracy with client-to-server update volume, server-to-client backbone synchronization, update freshness, and delayed-update handling. This framing is necessary because each policy changes not only the amount of communication, but also the timing and composition of the backbone updates that shape subsequent personalization.

\subsection{Accuracy and Communication under P1--P6}
\label{sec:p1-p6-results}

The P1--P6 comparison examines how gateway scheduling, offline-update handling, stale-update admission, and backbone-download control jointly determine decoding performance and communication under the default heterogeneous-link profile. Supplementary Table~S1 reports the complete operating points, including mean subject accuracy and the corresponding communication and synchronization measurements.

On OpenBMI, EIB-PH achieved higher mean subject accuracy than SB-PH under the ideal-link reference and every heterogeneous-link policy. Within EIB-PH, mean accuracy ranged from 80.13\% under P3 to 81.12\% under P5, compared with 80.75\% under the ideal-link reference. P5 also achieved the highest OpenBMI SB-PH accuracy, 73.02\%. P5 and P6 reduced server-to-client backbone traffic to 21.98~MB, compared with 37.49~MB under P3 and P4 and 54.56~MB under the ideal-link reference. These savings were accompanied by greater update rejection and accepted-update staleness, reflecting the trade-off introduced by communication-aware synchronization.

BCICIV-2a showed a different policy ordering. P4 achieved the highest mean accuracy under both SB-PH and EIB-PH, reaching 68.88\% and 69.80\%, respectively. Although P5 and P6 reduced server-to-client traffic to 3.68~MB, their accuracy depended more strongly on the personalization regime. P5 reached 68.44\% under EIB-PH and 66.77\% under SB-PH, while P6 reached 68.44\% and 64.51\%, respectively. The complete policy landscape therefore shows that communication efficiency and decoding accuracy cannot be ranked independently of the dataset and learning regime.

The full P1--P6 comparison uses a common training initialization and a matched gateway-availability trace to support controlled policy comparisons. Robustness of the principal P3/P5 comparison across five matched training and gateway-availability realizations under the default heterogeneous-link profile is summarized in Table~\ref{tab:p3p5-robustness}; complete replicate-level, subject-level, and communication results are reported in Supplementary Section~S2 (Tables~S2--S5).

Figure~\ref{fig:accuracy-communication} relates mean subject accuracy to server-to-client backbone traffic. Lower downlink traffic does not necessarily produce an accuracy-neutral operating point because scheduling, buffering, stale-update admission, and backbone-download decisions alter the sequence, recency, and composition of the updates that shape the shared backbone. The resulting points therefore represent coupled learning and synchronization outcomes rather than bandwidth reductions applied after training.

\begin{figure*}[t]
\centering
\includegraphics[width=0.98\textwidth]{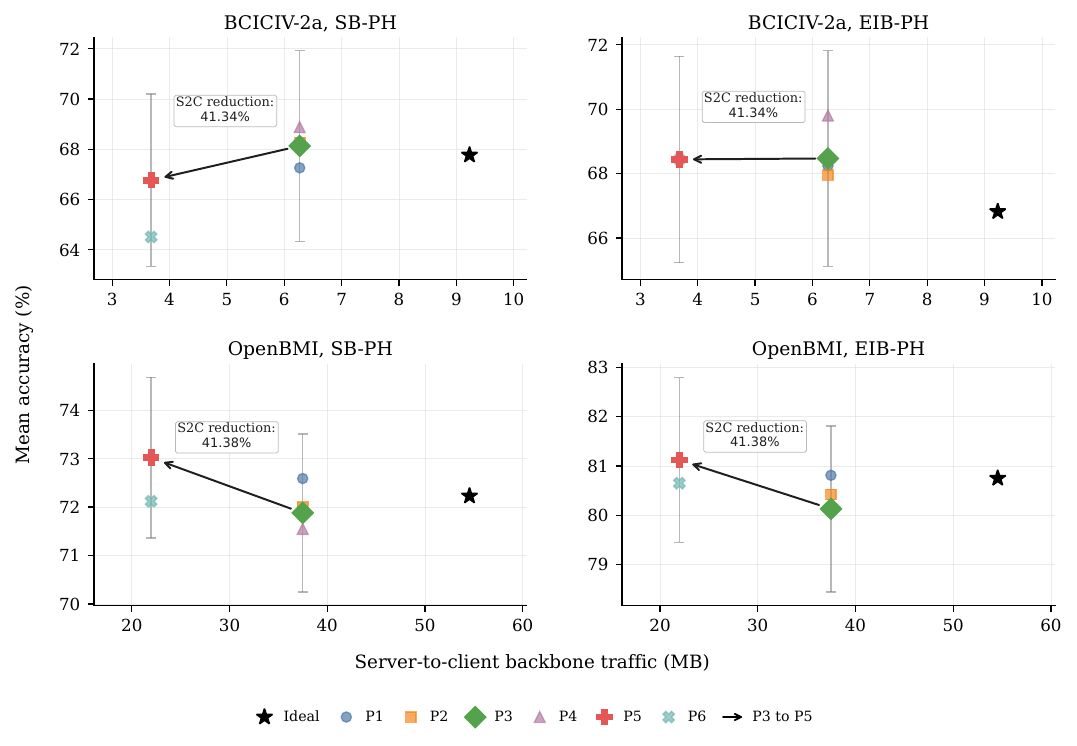}
\caption{Mean held-out decoding accuracy versus server-to-client (S2C) backbone traffic under the ideal-link reference and the default heterogeneous-link policies. For each policy, accuracy was first averaged within each subject across the three Session~2 calibration budgets ($k\in{15,20,30}$ trials per class) and then across subjects (BCICIV-2a, $n=9$; OpenBMI, $n=54$). The light-gray vertical error bars centered on P3 and P5 denote $\pm 1$ standard error across the subject-level averages. In each panel, the black arrow shows the change in the accuracy--downlink operating point from P3 to P5. The adjacent annotation reports P5's reduction in S2C backbone traffic relative to P3: 41.34\% for BCICIV-2a and 41.38\% for OpenBMI. Axis ranges are panel-specific; numerical values should be read from the axis ticks, and visual distances should not be compared across panels.}

\label{fig:accuracy-communication}
\end{figure*}

\begin{figure*}[t]
\centering
\includegraphics[width=0.98\textwidth]{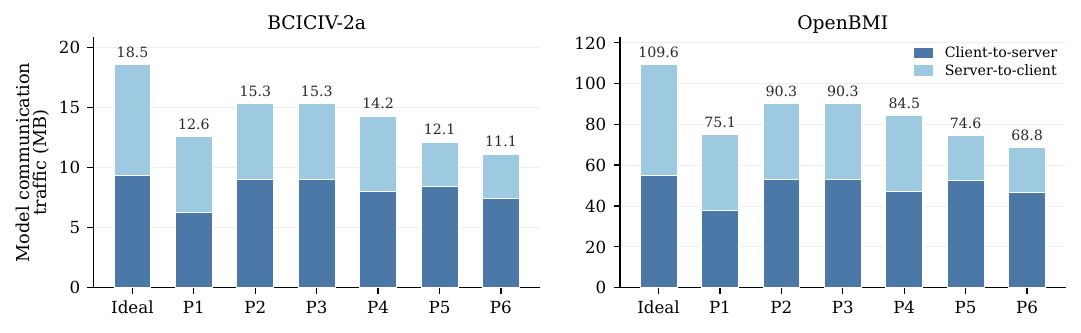}
\caption{Client-to-server (C2S) and server-to-client (S2C) model communication traffic for the ideal-link reference and the default heterogeneous-link policies. The plotted values are from EIB-PH; the corresponding SB-PH values are identical at the reported MB precision, indicating that the communication pattern shown is not specific to EIB-PH. The total bar height denotes aggregate model traffic, and stale-aware backbone downloading primarily reduces the S2C component.}
\label{fig:comm-components}
\end{figure*}

Figure~\ref{fig:comm-components} decomposes client-to-server and server-to-client traffic under EIB-PH. P5 and P6 produce the largest reduction in the server-to-client component, consistent with their stale-aware backbone-download rule. Client-to-server traffic varies less because fresh and buffered updates are still transmitted when gateways obtain upload opportunities.

\subsection{Effect of Communication-Aware Synchronization}
\label{sec:p3-p5-comparison}

The P3/P5 comparison is the central controlled analysis of communication-aware synchronization under heterogeneous gateway links. Both policies use FIFO buffering and stale-update rejection, so the offline-update and stale-admission rules are held fixed. They differ in two synchronization dimensions: gateway scheduling and backbone-download control. P3 uses non-adaptive all-gateway scheduling with always-download synchronization, whereas P5 combines communication-aware gateway scheduling with stale-aware backbone downloading. The comparison therefore isolates the joint effect of adapting gateway scheduling to realized availability and backbone refresh to version lag while holding the buffering and stale-update admission rules constant.

\begin{table*}[t]
\centering
\caption{Controlled P3/P5 comparison under the default heterogeneous-link condition.}
\label{tab:p3p5-controlled}

\begingroup
\scriptsize
\setlength{\tabcolsep}{3pt}
\renewcommand{\arraystretch}{1.06}

\resizebox{\textwidth}{!}{%
\begin{tabular}{llcrrrrrr}
\toprule
\multicolumn{9}{c}{\textbf{A. Observed operating points}} \\
\midrule

\textbf{Dataset} &
\textbf{Regime} &
\textbf{Policy} &
\makecell{\textbf{Mean subject}\\\textbf{accuracy (\%)}} &
\makecell{\textbf{C2S traffic}\\\textbf{(MB)}} &
\makecell{\textbf{S2C traffic}\\\textbf{(MB)}} &
\makecell{\textbf{Rejected uploads}\\\textbf{(\%)}} &
\makecell{\textbf{Accepted-update}\\\textbf{staleness (versions)}} &
\makecell{\textbf{Avoided downloads}\\\textbf{(\%)}} \\
\midrule

BCICIV-2a & SB-PH  & P3 & 68.13 & 9.02  & 6.27  & 19.20 & 0.27 & 0.00  \\
BCICIV-2a & SB-PH  & P5 & 66.77 & 8.44  & 3.68  & 24.59 & 0.52 & 35.92 \\
\addlinespace[2pt]

BCICIV-2a & EIB-PH & P3 & 68.47 & 9.02  & 6.27  & 19.20 & 0.27 & 0.00  \\
BCICIV-2a & EIB-PH & P5 & 68.44 & 8.44  & 3.68  & 24.59 & 0.52 & 35.92 \\
\midrule

OpenBMI & SB-PH  & P3 & 71.88 & 52.83 & 37.49 & 18.51 & 0.25 & 0.00  \\
OpenBMI & SB-PH  & P5 & 73.02 & 52.62 & 21.98 & 22.03 & 0.55 & 41.55 \\
\addlinespace[2pt]

OpenBMI & EIB-PH & P3 & 80.13 & 52.83 & 37.49 & 18.51 & 0.25 & 0.00  \\
OpenBMI & EIB-PH & P5 & 81.12 & 52.62 & 21.98 & 22.03 & 0.55 & 41.55 \\
\midrule

\multicolumn{9}{c}{\textbf{B. Paired accuracy effects and communication savings}} \\
\midrule

\textbf{Dataset} &
\textbf{Regime} &
\textbf{$n$} &
\makecell{\textbf{Paired difference}\\\textbf{(P5--P3, pp)}} &
\makecell{\textbf{95\% CI}\\\textbf{(pp)}} &
\makecell{\textbf{Holm-adjusted}\\\textbf{$p$-value}} &
\makecell{\textbf{S2C traffic}\\\textbf{saved (MB)}} &
\makecell{\textbf{S2C reduction}\\\textbf{(\%)}} &
\makecell{\textbf{Total traffic}\\\textbf{saved (MB)}} \\
\midrule

BCICIV-2a & SB-PH  & 9  & $-1.36$ & $[-5.35,\,2.62]$ & 0.906 & 2.59  & 41.34 & 3.17  \\
BCICIV-2a & EIB-PH & 9  & $-0.03$ & $[-2.47,\,2.42]$ & 0.981 & 2.59  & 41.34 & 3.17  \\
\midrule

OpenBMI & SB-PH  & 54 & $+1.14$ & $[0.14,\,2.14]$ & 0.078 & 15.51 & 41.38 & 15.72 \\
OpenBMI & EIB-PH & 54 & $+0.99$ & $[0.34,\,1.63]$ & 0.013 & 15.51 & 41.38 & 15.72 \\
\bottomrule
\end{tabular}%
}

\vspace{0.4em}
\begin{minipage}{0.98\textwidth}
\footnotesize
\emph{Notes:} Paired accuracy differences are computed as P5 minus P3. The 95\% CIs are two-sided Student-$t$ intervals based on paired subject-level differences. Holm adjustment is applied to the four paired $t$-test $p$-values defined by the two datasets and two learning regimes. S2C traffic saved and total traffic saved report the communication savings achieved by P5 relative to P3 and are computed as P3 minus P5; S2C reduction is likewise measured relative to P3. Total traffic is C2S plus S2C traffic. Rejected uploads are coordinator-rejected transmitted uploads; avoided downloads are normalized by synchronization opportunities. C2S and S2C denote client-to-server and server-to-client, respectively.
\end{minipage}

\endgroup
\end{table*}

Table~\ref{tab:p3p5-controlled} shows that P5 substantially reduces server-to-client backbone traffic relative to P3. On OpenBMI, server-to-client traffic decreases from 37.49~MB to 21.98~MB under both SB-PH and EIB-PH, corresponding to a 41.38\% reduction and 15.51~MB saved relative to P3. On BCICIV-2a, it decreases from 6.27~MB to 3.68~MB under both regimes, corresponding to a 41.34\% reduction and 2.59~MB saved relative to P3. When client-to-server and server-to-client traffic are combined, P5 saves 15.72~MB of total traffic on OpenBMI and 3.17~MB on BCICIV-2a relative to P3. By comparison, client-to-server traffic decreases by only 0.21~MB on OpenBMI and 0.58~MB on BCICIV-2a. The dominant communication saving therefore occurs in the server-to-client direction through P5's stale-aware backbone-download rule.

\begin{figure*}[t]
\centering
\includegraphics[width=0.98\textwidth]{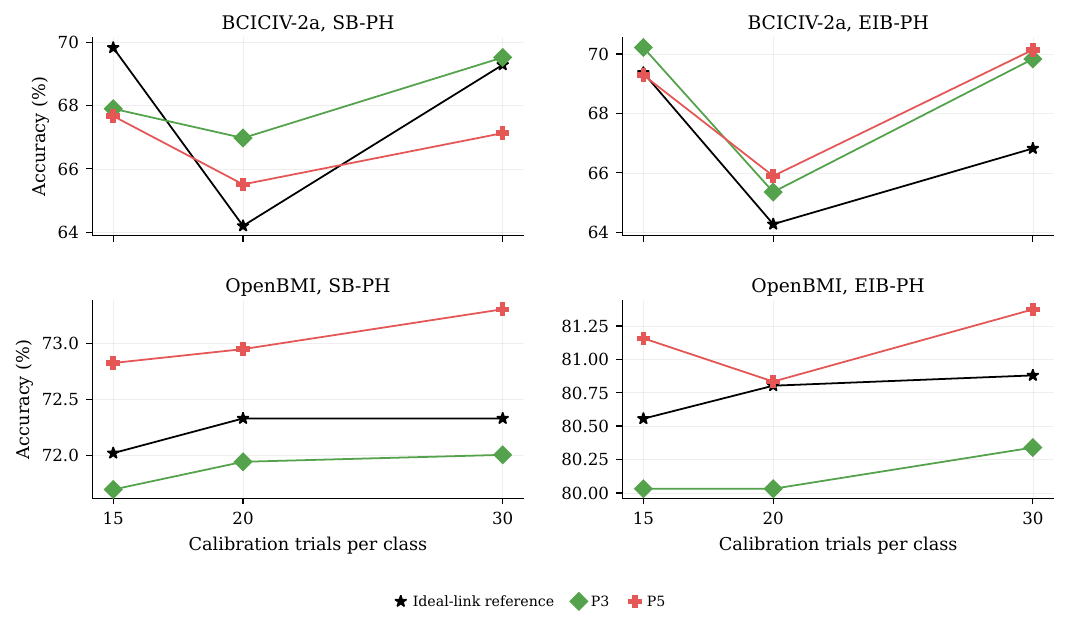}
\caption{Mean subject accuracy for P3 and P5 across Session~2 calibration budgets. The ideal-link result is included as the unconstrained synchronization reference for the corresponding dataset and learning regime.}
\label{fig:p3p5-calibration}
\end{figure*}

Figure~\ref{fig:p3p5-calibration} complements the aggregate comparison by showing P3 and P5 across the evaluated Session~2 calibration budgets, with the ideal-link result included as the unconstrained synchronization reference. On OpenBMI, P5 remains above P3 across all three calibration budgets under both learning regimes, with the strongest aggregate result under EIB-PH. On BCICIV-2a, the P3/P5 relationship varies across calibration budgets and learning regimes. The server-to-client communication reduction is therefore consistent, whereas the corresponding accuracy effect depends on the dataset and learning regime.

\begin{table*}[t]
\centering
\caption{Subject-level paired accuracy comparison of P5 and P3 under the default heterogeneous-link condition.}
\label{tab:p5p3-statistics}
\begingroup
\scriptsize
\setlength{\tabcolsep}{4pt}
\renewcommand{\arraystretch}{1.06}
\resizebox{\textwidth}{!}{%
\begin{tabular}{llrrrrrrr}
\toprule
\textbf{Dataset} &
\textbf{Regime} &
\textbf{$n$} &
\makecell{\textbf{Mean paired}\\\textbf{difference (pp)}} &
\makecell{\textbf{95\% Student-$t$ CI}\\\textbf{(pp)}} &
\makecell{\textbf{Paired effect}\\\textbf{size ($d_z$)}} &
\makecell{\textbf{Paired $t$-test}\\\textbf{$p$-value}} &
\makecell{\textbf{Wilcoxon signed-rank}\\\textbf{$p$-value}} &
\makecell{\textbf{Holm-adjusted}\\\textbf{$t$-test $p$-value}} \\
\midrule
BCICIV-2a & SB-PH  & 9  & -1.36 & $[-5.35,\ 2.62]$ & -0.26 & 0.453 & 0.496 & 0.906 \\
BCICIV-2a & EIB-PH & 9  & -0.03 & $[-2.47,\ 2.42]$ & -0.01 & 0.981 & 1.000 & 0.981 \\
\midrule
OpenBMI & SB-PH  & 54 & 1.14 & $[0.14,\ 2.14]$ & 0.31 & 0.026 & 0.016 & 0.078 \\
OpenBMI & EIB-PH & 54 & 0.99 & $[0.34,\ 1.63]$ & 0.42 & 0.003 & 0.001 & 0.013 \\
\bottomrule
\end{tabular}%
}

\vspace{0.4em}
\begin{minipage}{0.98\textwidth}
\footnotesize
\emph{Notes:} Accuracy differences are computed as P5 minus P3 and reported in percentage points. The 95\% CIs are two-sided Student-$t$ intervals based on paired subject-level accuracy differences. Holm adjustment is applied across the four paired $t$-tests defined by the two datasets and two learning regimes; Wilcoxon signed-rank tests are complementary unadjusted analyses.
\end{minipage}

\endgroup
\end{table*}

The strongest accuracy result occurs on OpenBMI under EIB-PH. In this setting, P5 increases mean subject accuracy from 80.13\% to 81.12\% relative to P3. The paired comparison yields a mean difference of +0.99 percentage points, with a 95\% confidence interval of [0.34, 1.63] percentage points and Holm-adjusted $p=0.013$. Table~\ref{tab:p3p5-controlled} summarizes this accuracy--communication result, while Table~\ref {tab:p5p3-statistics} reports the full paired statistical analysis. This is the clearest case in which communication-aware synchronization improves personalized accuracy while also reducing server-to-client backbone traffic.

Under OpenBMI SB-PH, P5 increases mean subject accuracy from 71.88\% to 73.02\%, with a mean paired difference of +1.14 percentage points. The unadjusted paired $t$-test gives $p=0.026$, but the Holm-adjusted value is $p=0.078$, so the improvement does not remain statistically significant after adjustment. On BCICIV-2a, P5 preserves the server-to-client communication reduction, but the accuracy differences are not statistically supported. Under EIB-PH, P5 and P3 are nearly identical at 68.44\% and 68.47\%, respectively, with Holm-adjusted $p=0.981$. Under SB-PH, P5 is lower than P3, at 66.77\% versus 68.13\%, but the paired confidence interval includes zero, and the Holm-adjusted value is $p=0.906$.

Taken together, the primary P3/P5 comparison shows that P5 reduced server-to-client backbone traffic by approximately 41\% on both datasets and also reduced total communication, with the only Holm-adjusted accuracy difference observed on OpenBMI under EIB-PH.

To assess whether this operating point depended on the primary controlled realization, the P3/P5 comparison was repeated across five matched training and gateway-availability realizations under the default heterogeneous-link profile. Table~\ref{tab:p3p5-robustness} summarizes the principal repeated-analysis outcomes; complete replicate-level, subject-level, and communication results are reported in Supplementary Section~S2 (Tables~S2--S5).

\begin{table}[t]
\centering
\caption{Robustness of the P3/P5 comparison across five matched training and gateway-availability realizations under the default heterogeneous-link profile.}
\label{tab:p3p5-robustness}
\begingroup
\scriptsize
\setlength{\tabcolsep}{4pt}
\renewcommand{\arraystretch}{1.06}

\begin{tabular*}{\columnwidth}{@{\extracolsep{\fill}}lcc@{}}
\toprule
\makecell[l]{\textbf{Condition}} &
\makecell[c]{\textbf{Mean accuracy difference}\\\textbf{(P5--P3, pp)}\\\textbf{[95\% CI]}} &
\makecell[c]{\textbf{Mean S2C reduction}\\\textbf{(\%)}\\\textbf{[observed range]}} \\
\midrule
BCICIV-2a, SB-PH  & $+0.80\;[-1.04,\,3.08]$ & $42.23\;[41.59,\,42.76]$ \\
BCICIV-2a, EIB-PH & $+0.88\;[-1.16,\,2.97]$ & $42.23\;[41.59,\,42.76]$ \\
\midrule
OpenBMI, SB-PH  & $-0.09\;[-0.78,\,0.67]$ & $41.98\;[41.47,\,42.47]$ \\
OpenBMI, EIB-PH & $-0.07\;[-0.53,\,0.39]$ & $41.98\;[41.47,\,42.47]$ \\
\bottomrule
\end{tabular*}

\vspace{0.4em}
\begin{minipage}{0.98\columnwidth}
\footnotesize
\emph{Notes:} Accuracy differences are computed as P5 minus P3 and reported in percentage points. The 95\% confidence intervals are obtained using a crossed hierarchical bootstrap across matched realizations and subjects. S2C reduction is measured relative to P3; the bracketed values report the observed minimum and maximum across the five realizations. S2C values are common to SB-PH and EIB-PH at the reported precision. S2C denotes server-to-client.
\end{minipage}

\endgroup
\end{table}

P5 reduced server-to-client backbone traffic in every matched realization, with mean reductions of 42.23\% on BCICIV-2a and 41.98\% on OpenBMI. In contrast, the accuracy differences were small and realization-dependent, and all four crossed hierarchical-bootstrap 95\% confidence intervals included zero. The repeated analysis therefore supports the robustness of the communication saving, but does not establish a consistent accuracy effect beyond the primary realization.

\subsection{Component Analysis of P5}
\label{sec:p5-component-analysis}

The controlled P3/P5 comparison establishes the net effect of the full communication-aware policy. However, P5 combines two coordination mechanisms that can influence the operating point in different ways: online-priority scheduling and stale-aware backbone downloading. We therefore decompose P5 on OpenBMI under EIB-PH, the setting with the largest subject cohort and the strongest controlled improvement of P5 relative to P3. This analysis separates online-first gateway selection from backbone-download control while keeping FIFO buffering and stale-update rejection fixed.

\begin{table*}[t]
\centering
\caption{Controlled component analysis of P5 on OpenBMI under EIB-PH and the default heterogeneous-link condition.}
\label{tab:p5-component-analysis}
\begingroup
\scriptsize
\setlength{\tabcolsep}{3pt}
\renewcommand{\arraystretch}{1.06}

\begin{tabular*}{\textwidth}{@{\extracolsep{\fill}}lll@{}}
\toprule
\multicolumn{3}{c}{\textbf{A. Controlled component variants}} \\
\midrule
\textbf{Variant} &
\textbf{Gateway-selection rule} &
\textbf{Backbone-download rule} \\
\midrule
P3 reference &
Fixed all-gateway selection &
Always download the current backbone \\

Online-random selection &
Random selection among currently online gateways &
Always download the current backbone \\

Online-priority selection &
\makecell[l]{Online-first priority selection based on version lag,\\
contribution recency, and pending-update state} &
Always download the current backbone \\

Stale-aware download only &
Fixed all-gateway selection &
Stale-aware backbone download \\

Full P5 &
\makecell[l]{Online-first priority selection based on version lag,\\
contribution recency, and pending-update state} &
Stale-aware backbone download \\
\bottomrule
\end{tabular*}

\vspace{0.6em}

\begin{tabular*}{\textwidth}{@{\extracolsep{\fill}}lrrrrrrr@{}}
\toprule
\multicolumn{8}{c}{\textbf{B. Accuracy and communication outcomes}} \\
\midrule
\textbf{Variant} &
\makecell{\textbf{Mean subject}\\\textbf{accuracy (\%)}} &
\makecell{\textbf{Accuracy difference}\\\textbf{from P3 (pp)}} &
\makecell{\textbf{C2S update}\\\textbf{traffic (MB)}} &
\makecell{\textbf{S2C backbone}\\\textbf{traffic (MB)}} &
\makecell{\textbf{S2C reduction}\\\textbf{from P3 (\%)}} &
\makecell{\textbf{Coordinator-rejected}\\\textbf{uploads (\%)}} &
\makecell{\textbf{Mean accepted-update}\\\textbf{staleness (versions)}} \\
\midrule
P3 reference &
80.13 & 0.00 & 52.83 & 37.49 & 0.00 & 18.51 & 0.25 \\

Online-random selection &
80.37 & +0.23 & 52.61 & 37.27 & 0.56 & 18.66 & 0.25 \\

Online-priority selection &
80.67 & +0.53 & 52.62 & 37.27 & 0.56 & 18.58 & 0.25 \\

Stale-aware download only &
80.43 & +0.30 & 52.83 & 21.98 & 41.38 & 22.01 & 0.55 \\

Full P5 &
81.12 & +0.99 & 52.62 & 21.98 & 41.38 & 22.03 & 0.55 \\
\bottomrule
\end{tabular*}

\vspace{0.4em}
\begin{minipage}{0.98\textwidth}
\footnotesize
\emph{Notes:} Panel A defines the controlled component variants, and Panel B reports their corresponding accuracy and communication outcomes. FIFO buffering and stale-drop admission are held fixed across all variants so that the effects of gateway selection and backbone-download control can be compared. Accuracy differences and S2C traffic reductions are reported relative to the P3 reference. C2S and S2C denote client-to-server and server-to-client, respectively.
\end{minipage}

\endgroup
\end{table*}

Table~\ref{tab:p5-component-analysis} shows that stale-aware backbone downloading produces the principal server-to-client traffic reduction. Adding stale-aware downloading without online-priority scheduling reduces server-to-client backbone traffic from 37.49~MB to 21.98~MB, matching the 41.38\% reduction obtained by full P5. In contrast, online-random and online-priority scheduling retain server-to-client traffic near the P3 reference, with reductions of only 0.56\%. The downlink saving observed under P5 is therefore primarily attributable to its backbone-download rule.

The accuracy results show a complementary contribution from gateway scheduling. Online-random scheduling increases mean subject accuracy from 80.13\% to 80.37\%, whereas online-priority scheduling reaches 80.67\%. The higher accuracy under online-priority scheduling indicates that prioritizing gateways according to synchronization need contributes beyond restricting participation to gateways that are currently online. Stale-aware downloading alone reaches 80.43\%, combining the full downlink reduction with a smaller accuracy increase. Full P5 attains the strongest component setting, reaching 81.12\% while retaining the same 21.98~MB of server-to-client backbone traffic as the stale-aware-download-only variant.

These results clarify the operating point achieved by P5 on OpenBMI under EIB-PH. Stale-aware backbone downloading controls server-to-client synchronization cost, while online-priority scheduling changes which available gateway updates shape the shared backbone. Their combination yields the highest observed accuracy among the component variants while preserving the full downlink reduction, reinforcing the conclusion that communication-aware coordination influences both synchronization cost and the learning trajectory of personalized MI-BCI.

\subsection{Sensitivity to Heterogeneous Link Availability}
\label{sec:link-sensitivity}

Gateway availability can change with wireless conditions, user mobility, device power state, and network congestion. We therefore test whether the P3/P5 conclusions remain stable as link availability changes from mild to default to severe conditions. The sensitivity analysis uses EIB-PH and compares P3 with P5 under three online-probability profiles: mild $0.98/0.85/0.60$, default $0.95/0.70/0.40$, and severe $0.90/0.50/0.20$ for high-, moderate-, and low-availability gateways. P3 and P5 are used because they share FIFO buffering and stale-update rejection, allowing the analysis to vary gateway scheduling and backbone-download control while holding delayed-update handling fixed.

\begin{figure*}[t]
\centering
\includegraphics[width=0.98\textwidth]{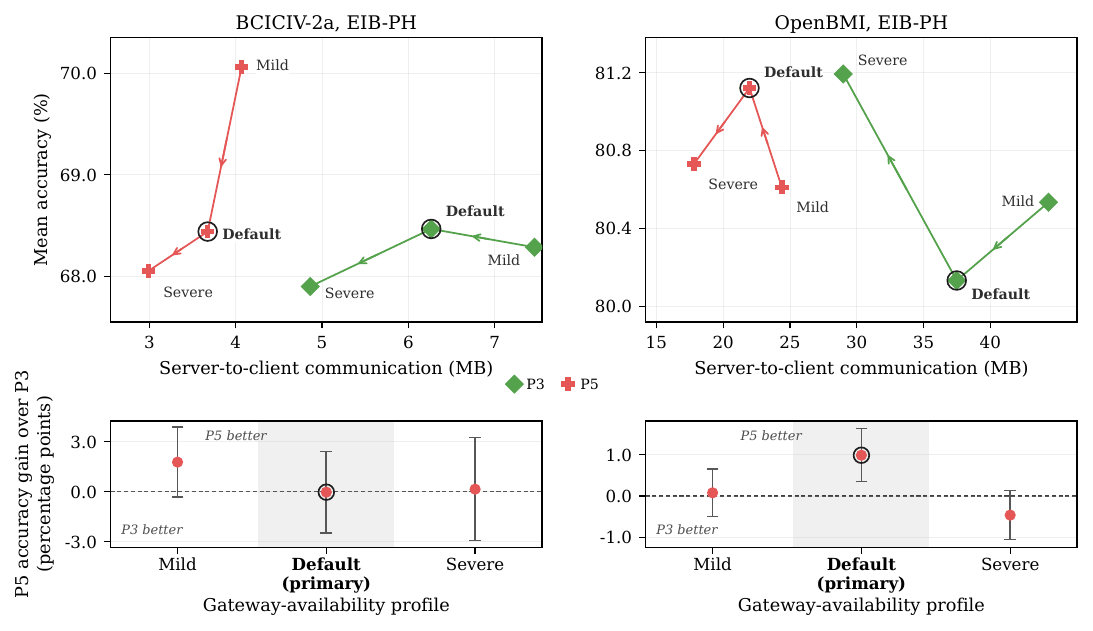}
\caption{Sensitivity of the controlled P3/P5 comparison to mild, default, and severe heterogeneous gateway availability under EIB-PH. Both policies use FIFO buffering and stale-update rejection, isolating the combined effect of communication-aware scheduling and stale-aware backbone downloading. Accuracy is averaged within each subject across $k\in{15,20,30}$ calibration trials per class and then across subjects (BCICIV-2a, $n=9$; OpenBMI, $n=54$). The upper panels show mean held-out accuracy versus server-to-client backbone traffic, with arrows tracing mild to default to severe availability; P5 uses less traffic under every profile. The lower panels show mean paired P5-minus-P3 accuracy differences with two-sided Student's $t$ 95\% confidence intervals; positive values favor P5, negative values favor P3, and the dashed line denotes no difference. The light-gray bands behind the lower-panel default columns and the black rings around default markers in both rows identify the prespecified primary profile. Only the OpenBMI default-profile improvement is statistically supported (Holm-adjusted $p=0.013$); the remaining comparisons are not statistically resolved.}

\label{fig:link-sensitivity}
\end{figure*}

Figure~\ref{fig:link-sensitivity} summarizes the principal accuracy--communication trends across the availability profiles, while Supplementary Table~S6 reports the complete accuracy, communication, and synchronization results. Coordinator-rejected upload rates and buffered-upload delays increase as the profile moves from mild to severe. On OpenBMI, the coordinator-rejected upload rate under P3 rises from 8.63\% to 27.52\%, while mean buffered-upload delay increases from 1.33 to 1.60 rounds. Under P5, the corresponding values rise from 13.60\% to 30.64\% and from 1.33 to 1.59 rounds. Mean accepted-update staleness remains higher under P5 than under P3 across all three profiles, ranging from 0.51 to 0.55 backbone versions under P5 and from 0.21 to 0.26 backbone versions under P3. BCICIV-2a follows the same overall pattern: coordinator rejection and buffered-upload delay increase as availability becomes more constrained, while accepted-update staleness remains higher under P5. These results confirm that severe availability creates a more delayed and more selective update stream for both policies.

Across all availability profiles, P5 maintains a lower server-to-client synchronization cost than P3. On OpenBMI, P5 reduces server-to-client backbone traffic from 44.36~MB to 24.39~MB under mild availability, from 37.49~MB to 21.98~MB under the default profile, and from 28.99~MB to 17.81~MB under severe availability. On BCICIV-2a, the corresponding reductions are from 7.46~MB to 4.07~MB, from 6.27~MB to 3.68~MB, and from 4.86~MB to 2.99~MB. Thus, the downlink-saving effect of P5 persists across the full range of evaluated link conditions.

The accuracy effect is less stable than the communication effect. On OpenBMI, P5 is nearly tied with P3 under mild availability, 80.61\% versus 80.53\%; it is higher under the default profile, 81.12\% versus 80.13\%; and it is lower under severe availability, 80.73\% versus 81.19\%. The default-profile improvement corresponds to the prespecified primary comparison reported in Table~\ref{tab:p5p3-statistics} (Holm-adjusted $p=0.013$), while the 95\% confidence intervals for the mild- and severe-profile comparisons include zero. On BCICIV-2a, P5 is higher than P3 under mild availability, 70.06\% versus 68.29\%, and nearly matched under the default and severe profiles; the 95\% confidence intervals for all three profile-specific comparisons include zero.

These results refine the P3/P5 conclusion. P5 consistently reduces server-to-client backbone synchronization as link availability changes, but its accuracy effect depends on the availability profile and dataset. More constrained availability also increases coordinator-rejected uploads and buffered-upload delay, while accepted-update staleness remains dependent on the policy and availability profile. The same policy can therefore operate on a more delayed and more selective update stream as connectivity deteriorates. Thus, communication-aware synchronization should not be treated as a universal default without validation on the intended user cohort, personalization regime, and gateway-availability conditions.

\subsection{Subject-Level Reliability}
\label{sec:subject-reliability}

Aggregate accuracy is necessary but insufficient for MI-BCI deployment. A communication policy can improve the cohort mean while leaving some users with degraded or low-accuracy personalized decoders. We therefore compare each subject's heterogeneous-link accuracy with that subject's matched ideal-link reference under EIB-PH and summarize the resulting reliability profiles in Table~\ref{tab:subject-reliability}. For threshold-based reporting, we use 60\% accuracy for the four-class BCICIV-2a task and 70\% accuracy for the binary OpenBMI task. These thresholds are analytical markers rather than clinical usability criteria. They identify low-accuracy operating regions while accounting for the different class structures and expected performance ranges of the two tasks.

\begin{table*}[t]
\centering
\caption{Subject-level reliability under EIB-PH relative to the ideal-link reference.}
\label{tab:subject-reliability}
\begingroup
\scriptsize
\setlength{\tabcolsep}{3pt}
\renewcommand{\arraystretch}{1.06}

\begin{tabular*}{\textwidth}{@{\extracolsep{\fill}}llrrrrr@{}}
\toprule
\multicolumn{7}{c}{\textbf{A. Subject-level accuracy differences relative to the ideal-link reference}} \\
\midrule
\textbf{Dataset} &
\textbf{Policy} &
\makecell{\textbf{Mean accuracy}\\\textbf{difference (pp)}} &
\makecell{\textbf{Median accuracy}\\\textbf{difference (pp)}} &
\makecell{\textbf{Minimum subject}\\\textbf{difference (pp)}} &
\makecell{\textbf{Maximum subject}\\\textbf{difference (pp)}} &
\makecell{\textbf{Difference range}\\\textbf{(max--min, pp)}} \\
\midrule
BCICIV-2a & P1 &  1.41 &  1.62 &  -6.71 &  7.18 & 13.89 \\
BCICIV-2a & P2 &  1.13 &  1.39 &  -3.47 &  6.48 &  9.95 \\
BCICIV-2a & P3 &  1.65 &  3.94 & -14.12 & 10.42 & 24.54 \\
BCICIV-2a & P4 &  2.98 &  5.56 & -14.12 &  8.33 & 22.45 \\
BCICIV-2a & P5 &  1.62 &  3.47 & -12.04 &  9.26 & 21.30 \\
BCICIV-2a & P6 &  1.62 &  3.47 &  -5.56 &  5.79 & 11.34 \\
\midrule
OpenBMI & P1 &  0.06 &  0.00 & -6.67 & 5.83 & 12.50 \\
OpenBMI & P2 & -0.32 &  0.00 & -7.22 & 5.00 & 12.22 \\
OpenBMI & P3 & -0.61 & -0.28 & -8.61 & 3.06 & 11.67 \\
OpenBMI & P4 & -0.53 & -0.56 & -9.17 & 4.17 & 13.33 \\
OpenBMI & P5 &  0.38 &  0.28 & -7.22 & 5.83 & 13.06 \\
OpenBMI & P6 & -0.10 & -0.14 & -5.83 & 5.28 & 11.11 \\
\bottomrule
\end{tabular*}

\vspace{0.6em}

\begin{tabular*}{\textwidth}{@{\extracolsep{\fill}}llrrrrr@{}}
\toprule
\multicolumn{7}{c}{\textbf{B. Absolute accuracy and subject-level reliability counts}} \\
\midrule
\textbf{Dataset} &
\textbf{Policy} &
\makecell{\textbf{Minimum subject}\\\textbf{accuracy (\%)}} &
\makecell{\textbf{Subjects below}\\\textbf{threshold ($n$)}} &
\makecell{\textbf{Subjects with $\geq$5-pp}\\\textbf{accuracy decrease ($n$)}} &
\makecell{\textbf{Subjects with $\geq$5-pp}\\\textbf{accuracy increase ($n$)}} &
\makecell{\textbf{Below threshold with}\\\textbf{$\geq$5-pp decrease ($n$)}} \\
\midrule
BCICIV-2a & P1 & 54.86 & 3 & 1 & 2 & 0 \\
BCICIV-2a & P2 & 49.54 & 3 & 0 & 1 & 0 \\
BCICIV-2a & P3 & 59.49 & 3 & 1 & 3 & 0 \\
BCICIV-2a & P4 & 57.41 & 1 & 1 & 5 & 0 \\
BCICIV-2a & P5 & 54.17 & 2 & 1 & 2 & 0 \\
BCICIV-2a & P6 & 52.08 & 3 & 1 & 2 & 0 \\
\midrule
OpenBMI & P1 & 53.06 & 12 & 2 & 1 & 1 \\
OpenBMI & P2 & 53.61 & 10 & 4 & 1 & 2 \\
OpenBMI & P3 & 52.50 & 10 & 3 & 0 & 1 \\
OpenBMI & P4 & 53.61 & 12 & 3 & 0 & 3 \\
OpenBMI & P5 & 53.33 & 12 & 2 & 4 & 1 \\
OpenBMI & P6 & 53.06 & 12 & 1 & 1 & 1 \\
\bottomrule
\end{tabular*}

\vspace{0.4em}
\begin{minipage}{0.98\textwidth}
\footnotesize
\emph{Notes:} Panel A summarizes subject-level accuracy differences relative to the matched ideal-link reference, computed as policy minus ideal-link accuracy. Panel B reports absolute accuracy and subject-level reliability counts. The analytical low-accuracy thresholds are 60\% for BCICIV-2a and 70\% for OpenBMI; subjects exactly at the threshold are not counted as below threshold. Large decreases and increases include changes of exactly 5~pp. The joint count includes subjects who are both below the corresponding threshold and at least 5~pp below their matched ideal-link accuracy.
\end{minipage}

\endgroup
\end{table*}

OpenBMI illustrates why subject-level reliability must be examined alongside aggregate accuracy. Under EIB-PH, P5 has a mean subject-level accuracy difference of +0.38 percentage points relative to the ideal-link reference, whereas P3 has a mean difference of -0.61 percentage points. P5 also increases the number of subjects with an accuracy gain of at least 5~pp from zero under P3 to four. However, P5 still has two subjects with an accuracy decrease of at least 5~pp, a minimum subject-level difference of -7.22 percentage points, and 12 subjects below the 70\% analysis threshold. Only one subject under P5 is both below this threshold and degraded by at least 5~pp relative to the matched ideal-link reference. This distinction shows that low absolute accuracy and substantial degradation relative to the ideal-link reference capture related but distinct aspects of subject-level reliability. Its subject-level difference range is $13.06$ percentage points. Thus, the policy with the strongest aggregate OpenBMI result still leaves a nontrivial low-accuracy tail.

BCICIV-2a shows an even sharper separation between cohort-level improvement and subject-level reliability. Under EIB-PH, P4 has the largest mean subject-level difference, +2.98 percentage points, and five subjects improve by at least 5~pp. At the same time, its minimum subject-level difference is -14.12 percentage points, and its difference range is 22.45 percentage points. P5 exhibits a similar tail pattern, with a mean difference of +1.62 percentage points, a minimum difference of -12.04 percentage points, and a difference range of 21.30 percentage points. P6 has the same mean difference as P5, +1.62 percentage points, but a substantially narrower difference range of 11.34 percentage points and a minimum difference of -5.56 percentage points. Notably, under every P1--P6 policy, no BCICIV-2a subject was simultaneously below the 60\% analytical threshold and degraded by at least 5~pp relative to the matched ideal-link reference. In other words, for each policy, the subjects below the 60\% analytical threshold were not the same subjects as those who experienced a decline of at least 5~pp relative to the matched ideal-link reference. These results show that the policy with the strongest cohort mean is not necessarily the policy with the most favorable subject-level tail behavior.

\begin{figure*}[t]
\centering
\includegraphics[width=0.86\textwidth]{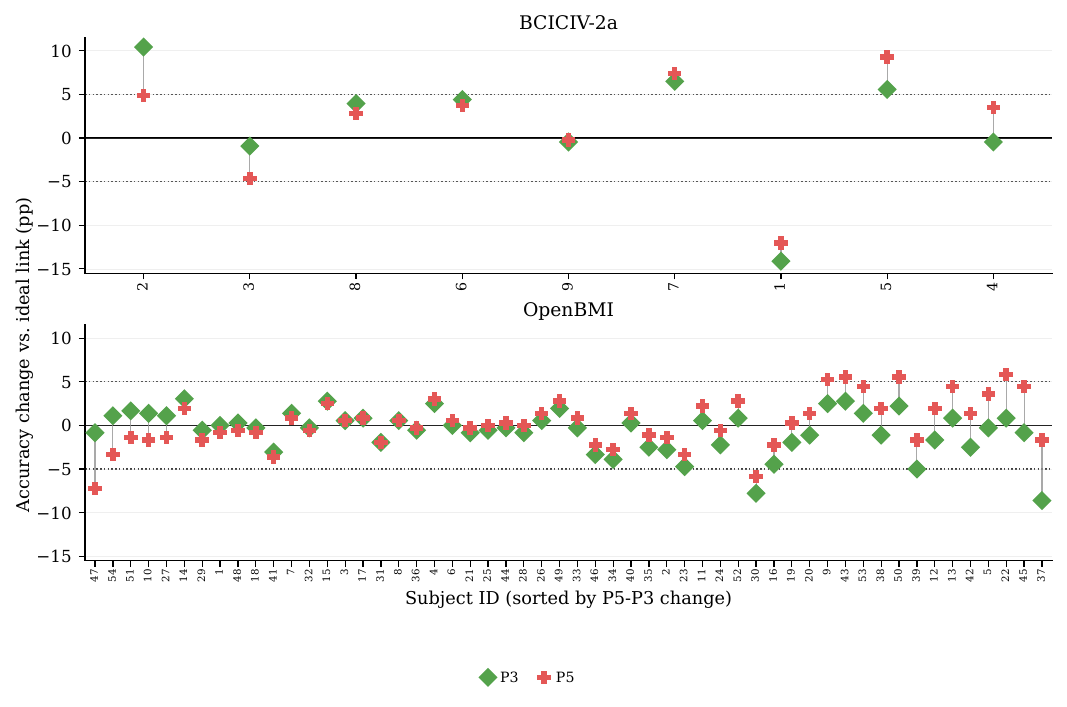}
\caption{Subject-level accuracy changes under EIB-PH relative to each subject's matched ideal-link reference, averaged across $k\in{15,20,30}$ calibration trials per class (BCICIV-2a, $n=9$; OpenBMI, $n=54$). Each vertical pair connects the P3 and P5 changes for one subject, with subjects sorted by the P5-minus-P3 difference; dotted lines mark changes of $\pm5$ percentage points. These values are relative changes, not absolute accuracies: a subject may improve relative to ideal link while still have low absolute accuracy. Table~\ref{tab:subject-reliability} reports absolute accuracy and counts below the dataset-specific analysis thresholds.}

\label{fig:subject-reliability}
\end{figure*}

We next examine whether subject-level accuracy differences covary with baseline decoding strength or communication exposure. Supplementary Table~S7 reports descriptive Spearman rank correlations together with 95\% bootstrap confidence intervals. On BCICIV-2a, ideal-link accuracy produced the largest point estimate under both P3 and P5 ($\rho=-0.68$), but the corresponding confidence intervals were wide and included zero. On OpenBMI, all point estimates were small, and all reported confidence intervals likewise included zero. Across datasets and policies, no recorded communication-exposure quantity showed a consistent association with subject-level accuracy differences. These exploratory correlations therefore do not support attributing subject-level vulnerability to rejected uploads, accepted-update staleness, buffering, gateway availability, or avoided backbone downloads individually.

These reliability results have direct significance for MI-BCI deployment. For users already operating in a low-accuracy range, an additional policy-induced decline can make motor-intent feedback unreliable and disrupt continued home neurorehabilitation. Operationally, a persistent threshold violation or a large decline from the subject's reference performance could prompt the collection of a small additional calibration set, a temporary shift to more conservative synchronization that favors fresher backbone states, or rollback to the most recent locally validated decoder state. Monitoring both cohort-level and subject-level outcomes is therefore essential in home use, where frequent clinician-supervised recalibration may be impractical.

\section{Discussion and Deployment Implications}
\label{sec:discussion}

The results position federated MI-BCI personalization as a deployment-coupled learning problem rather than a model-only problem. In a gateway-coordinated system, communication policy determines which subject updates reach the coordinator, when delayed updates are admitted, and which backbone version is available for subsequent local training and personalization. Synchronization behavior therefore influences the trajectory of the shared representation as well as the communication required to maintain it. The differences observed across P1--P6 show that gateway coordination cannot be treated as a bandwidth-control layer applied independently of learning.

This perspective extends prior federated EEG and MI-BCI research. Existing studies demonstrate that federated learning can exploit cross-subject information, address non-IID subject data, and support collaboration across heterogeneous EEG datasets without centralizing the underlying recordings \cite{jia2024federated, ju2020federated, liu2024aggregating}. Their principal emphasis is the learning method and its resulting classification performance under a prescribed federation process. NEXUS-MI addresses a complementary deployment question: whether later-session personalization remains effective when the shared representation is maintained through intermittent and policy-controlled gateway synchronization. It therefore makes gateway scheduling, offline buffering, stale-update admission, and backbone-download control explicit experimental variables and relates them to the personalized decoder ultimately available to each subject.

NEXUS-MI also differs in emphasis from communication-aware federated-learning systems such as Oort and FedBuff, which study participant selection and buffered asynchronous aggregation primarily through time-to-accuracy, scalability, convergence, and aggregate model utility \cite{lai2021oort, nguyen2022federated}. The present framework applies related coordination principles to MI-BCI, where communication efficiency must be interpreted together with limited-calibration personalization and subject-level reliability. The component analysis illustrates this distinction: stale-aware backbone downloading produces the principal reduction in server-to-client traffic, while online-priority scheduling changes which available subject updates shape the shared backbone. Their combination affects both synchronization cost and the subsequent personalization outcome.

The dataset-dependent findings further show why no single communication policy should be assumed to transfer uniformly across MI-BCI cohorts. OpenBMI provides a larger 54-subject cohort and a binary motor-imagery task, whereas BCICIV-2a contains nine subjects and requires four-class discrimination. The larger OpenBMI cohort provides a broader basis for pooled Session~1 representation initialization and collaborative backbone learning, which is consistent with the clear EIB-PH advantage and the comparatively stable P3/P5 behavior observed on that dataset. In BCICIV-2a, each subject contributes a larger share of the collaborative update stream, and the four-class personalized decision problem is more demanding. These characteristics are consistent with the greater subject-level sensitivity observed under constrained synchronization. More broadly, the contrast reflects the established dependence of EEG transfer and MI-BCI performance on subject, session, and task heterogeneity \cite{wu2020transfer, ahn2015performance}. Communication-policy conclusions must therefore be validated for the intended cohort, learning regime, and calibration protocol rather than inferred from a single benchmark.

The evaluation also shows that communication efficiency must be interpreted together with reliability. Reducing server-to-client backbone synchronization is valuable for wearable, edge-enabled, and home-oriented MI-BCI, where gateway connectivity and bandwidth may be limited \cite{arpaia2023paving,wang2020accurate}. However, lower traffic alone does not establish a suitable operating point. The approximately 41\% reduction observed in the primary P5-versus-P3 comparison remained approximately 42\% across five matched realizations, whereas the corresponding accuracy differences were small and realization-dependent. This stable communication saving must therefore be interpreted alongside the subject-level outcomes concealed by the cohort mean.

Importantly, this operating point is achieved without transferring raw EEG trials, calibration labels, or personalized classifier heads during the gateway-coordinated phase. These remain at the subject-side gateway, while collaboration is limited to backbone-related updates, refreshed backbone states, and synchronization metadata.

Subject-level reliability is particularly important for assistive and rehabilitation-oriented MI-BCI because the decoder determines the feedback experienced by each user. A cohort-level improvement does not compensate for a user whose personalized decoder remains inaccurate or deteriorates substantially under constrained synchronization. For users already operating in a low-accuracy range, an additional policy-induced decline can make motor-intent feedback unreliable and disrupt continued home neurorehabilitation \cite{ahn2015performance}. Operationally, a persistent threshold violation or a large decline from the subject's reference performance could prompt the collection of a small additional calibration set, a temporary shift to more conservative synchronization that favors fresher backbone states, a user-specific backbone refresh, or rollback to the most recent locally validated decoder state. These responses are especially relevant in home use, where frequent clinician-supervised recalibration may be impractical \cite{arpaia2023paving}.

\begin{table*}[t]
\centering
\caption{Design guidance for edge-enabled MI-BCI deployment.}
\label{tab:deployment-implications}
\begingroup
\footnotesize
\renewcommand{\arraystretch}{1.10}
\begin{tabular}{p{0.40\textwidth}p{0.54\textwidth}}
\toprule
\textbf{Empirical lesson} & \textbf{Deployment guidance} \\
\midrule
A constrained-link result is meaningful only when interpreted relative to an ideal-link reference for the same learning protocol. & Establish an unconstrained synchronization reference before deploying communication controls. Report how much accuracy, subject-level reliability, and synchronization cost are retained when gateway links become intermittent or selective. \\

Communication policy changes the learning process, not only the number of transmitted bytes. & Treat gateway coordination as a control plane for learning. Client selection, buffering, stale-update admission, and backbone-refresh timing should be tuned jointly rather than optimized as independent communication knobs. \\

Server-to-client backbone synchronization is a major controllable cost in gateway-coordinated personalization. & Avoid unconditional backbone downloads. Version-aware and staleness-aware refresh rules are practical mechanisms for reducing edge-to-gateway traffic while allowing gateways to continue local personalization from cached model states. \\

The accuracy effect of communication-aware synchronization is not uniform across datasets, regimes, and calibration budgets. & Do not deploy a bandwidth-saving scheduler as a universal default. Accept a communication policy only after validating that it preserves cohort-level accuracy and does not introduce unacceptable subject-level degradation on the target population. \\

Aggregate accuracy can mask individual users who experience large losses or remain below a useful operating threshold. & Monitor per-user reliability in addition to cohort means. Worst-subject accuracy, threshold violations, large reference-relative declines, and cases where low accuracy coincides with a large decline should inform conservative synchronization, additional calibration, or rollback. \\

Dropped uploads, update staleness, buffering, and avoided downloads are useful exposure signals, but none showed a consistent association with subject-level accuracy differences. & Instrument gateways and the edge coordinator to log communication exposure together with decoding outcomes. Use these measures to support operational monitoring, but not as standalone explanations or predictors of subject-level vulnerability. \\

Collaborative personalization can maintain raw-EEG data locality. & Keep raw EEG and subject-specific classifier heads at the gateway; exchange only backbone-related parameters and synchronization metadata. \\

\bottomrule
\end{tabular}

\vspace{2pt}
\parbox{0.96\textwidth}{%
\textit{Note:} The empirical observations summarized here are limited to the evaluated BCICIV-2a and OpenBMI datasets and the modeled heterogeneous gateway-to-edge communication conditions.}

\endgroup
\end{table*}

Table~\ref{tab:deployment-implications} summarizes the resulting design guidance. The broader implication is that future MI-BCI federated learning studies should report communication behavior and per-user reliability alongside decoding accuracy. Client-to-server traffic, server-to-client synchronization, update rejection, staleness, buffering, and download avoidance describe the system-level operating point. Subject-level deviations, threshold violations, the co-occurrence of low absolute accuracy and large degradation from the ideal-link reference, and worst-subject behavior describe whether that operating point remains reliable for individual users. Reporting these dimensions together provides a more deployment-relevant standard for evaluating federated personalization in MI-BCI.

The scope of these findings is limited to an offline, replay-based evaluation using two public session-based MI-BCI datasets and a one-subject/one-gateway abstraction. Both BCICIV-2a and OpenBMI comprise recordings from healthy participants collected under controlled laboratory conditions; the present evaluation therefore does not establish performance for stroke survivors or other rehabilitation populations. Although this design supports controlled analysis of gateway synchronization, it does not capture the behavioral adaptation, longitudinal EEG variability, clinician interaction, or network dynamics of a live wearable system used in the home. Connectivity is represented through fixed Bernoulli availability profiles and fixed policy parameters, which provide reproducible operating conditions but do not reproduce bursty outages, correlated link failures, or time-varying bandwidth in operational networks.

The principal P3/P5 comparison was repeated across five matched training and gateway-availability realizations, showing that the approximately 42\% reduction in server-to-client backbone traffic persists beyond the primary controlled realization. The associated accuracy differences were small and realization-dependent, and all crossed hierarchical-bootstrap 95\% confidence intervals included zero. The repeated analysis was limited to P3 and P5 under the default availability profile; the broader P1--P6 policy landscape, component analysis, availability-severity study, and subject-level association analysis remain based on the primary realization. Because several stochastic elements of the experimental realization vary across the matched replicates, the repeated analysis evaluates their combined influence rather than isolating individual sources of run-to-run variability. Although NEXUS-MI reduced model-payload traffic, the experiments did not evaluate its end-to-end deployment cost. Under the evaluated configuration of 120 communication rounds and 50 local epochs per generated update, gateway computation time, wall-clock synchronization latency, energy consumption, and network-protocol overhead were not measured. The reported reductions therefore demonstrate model-payload communication savings under controlled offline replay, but do not yet establish real-time or energy-efficient operation on physical gateways. Future work should profile gateway computation time, wall-clock synchronization latency, energy consumption, and complete network overhead on representative gateway hardware. It should also extend the repeated evaluation to the full policy space and broader connectivity conditions, incorporate measured traces from representative home networks, and validate the framework longitudinally in rehabilitation populations. Adaptive per-user synchronization is another important direction, with gateway scheduling, backbone-refresh frequency, and stale-update handling responding to each user's recent decoder stability, calibration need, and observed link condition rather than remaining fixed across the cohort.

\section{Conclusion}
\label{sec:conclusion}

This study shows that gateway synchronization is not merely a communication-layer concern in federated motor-imagery BCI personalization; it helps determine the shared representation from which each subject's decoder is personalized. Across the six evaluated policies, no strategy is uniformly dominant across datasets and personalization regimes. The principal controlled comparison nevertheless demonstrates a consistent systems benefit: communication-aware coordination reduces server-to-client backbone traffic by approximately \(42\%\) across five matched realizations. The corresponding accuracy effect is not reproducible, however, and cohort averages conceal subject-level losses and low-accuracy cases. Improved communication efficiency should therefore not be interpreted as evidence that personalized decoding performance is preserved for every user.

These findings support a deployment-centered evaluation approach in which personalized accuracy is considered jointly with bidirectional communication, update freshness, link-availability sensitivity, and subject-level reliability. A suitable coordination strategy is not simply the one that minimizes traffic; it must also maintain acceptable performance for individual users as connectivity and EEG conditions vary.

The evidence remains limited to offline replay of public laboratory datasets, predefined Bernoulli gateway-availability profiles with fixed policy parameters, and five matched realizations of the principal comparison. Future work should validate NEXUS-MI longitudinally under measured network conditions and evolving EEG distributions, including bursty and correlated outages. It should also develop user-specific synchronization that responds jointly to availability, version lag, calibration need, and decoder stability, with safeguards such as additional calibration or rollback when reliability deteriorates. These steps would move communication-aware federated personalization toward dependable edge-enabled and home-oriented deployment.

\section*{Author Contributions}
Daniel Adu Worae: Conceptualization; Methodology; Software; Investigation; Formal analysis; Writing - original draft; Writing - review \& editing.

Aarthy Nagarajan: Conceptualization; Methodology; Supervision; Writing - review \& editing.

\section*{Funding}
This work received institutional research support associated with the Melchor Visiting Assistant Professorship and the Lucy Family Institute for Data and Society at the University of Notre Dame. No specific external grant supported this work.

\section*{Acknowledgments}
The authors thank Professor Nitesh Chawla for his valuable feedback on this work.

\section*{Ethics Statement}
The present study analyzed existing publicly available, de-identified EEG data from BCICIV-2a and OpenBMI. No participants were recruited and no new human-subject data were collected for this secondary analysis. No additional institutional ethics review was required for this secondary analysis of publicly available, de-identified data. For OpenBMI, the original study was reviewed and approved by the Korea University Institutional Review Board (1040548-KUIRB-16-159-A-2), and written informed consent was obtained from all participants \cite{lee2019eeg}. BCICIV-2a was analyzed as the publicly released, de-identified BCI Competition IV dataset; the public dataset description used for this secondary analysis does not provide an ethics-committee reference number \cite{bciciv2a_dataset,tangermann2012review}. The manuscript contains no identifiable participant information, images, or videos.

\section*{Declaration of Competing Interest}
The authors declare that they have no known competing financial interests or personal relationships that could have appeared to influence the work reported in this paper.

\section*{Data and Code Availability}
The datasets analyzed in this study are publicly available from their original repositories. BCICIV-2a is available through the BCI Competition IV and BNCI Horizon 2020 repositories as Dataset 001-2014 \cite{bciciv2a_dataset}, and OpenBMI is available through GigaDB under DOI 10.5524/100542 \cite{openbmi_dataset}. The NEXUS-MI code, experiment outputs, and supporting reproducibility materials are publicly available at \url{https://github.com/WadElla/NEXUS-MI} and are permanently archived on Zenodo under DOI 10.5281/zenodo.22074301 \cite{worae2026nexusmi}. The reference environment used for the repeated experimental runs was Linux 6.8 (glibc 2.35) with Python 3.10.19, PyTorch 2.7.1, CUDA 11.8, NumPy 2.2.5, and an NVIDIA CUDA-capable device. The archived software supports Python 3.10 or later, and complete dependency and execution requirements are documented with the software release. NEXUS-MI is distributed under the MIT License.

\section*{Declaration of Generative AI and AI-Assisted Technologies in Manuscript Preparation}
During the preparation of this work, the authors used ChatGPT (OpenAI) to assist with language refinement, code development, debugging, and statistical analysis of author-generated results. The authors reviewed and edited all AI-assisted outputs, and they tested and validated the code before use. The authors take full responsibility for the content of the publication.

\balance
\bibliographystyle{IEEEtran}
\bibliography{references}

\clearpage
\setcounter{section}{0}
\setcounter{subsection}{0}
\setcounter{table}{0}
\setcounter{figure}{0}
\setcounter{equation}{0}
\renewcommand{\thesection}{S\arabic{section}}
\renewcommand{\thesubsection}{\thesection.\Alph{subsection}}
\renewcommand{\thetable}{S\arabic{table}}
\renewcommand{\thefigure}{S\arabic{figure}}
\renewcommand{\theequation}{S\arabic{equation}}

\twocolumn[{
\centering
{\large\scshape Supplementary Material\par}
\vspace{0.8em}
}]

\section{Complete Policy Operating Points}
\label{sec:supp-policy-landscape}

\begin{figure*}[!b]
\centering
\includegraphics[width=0.90\textwidth]{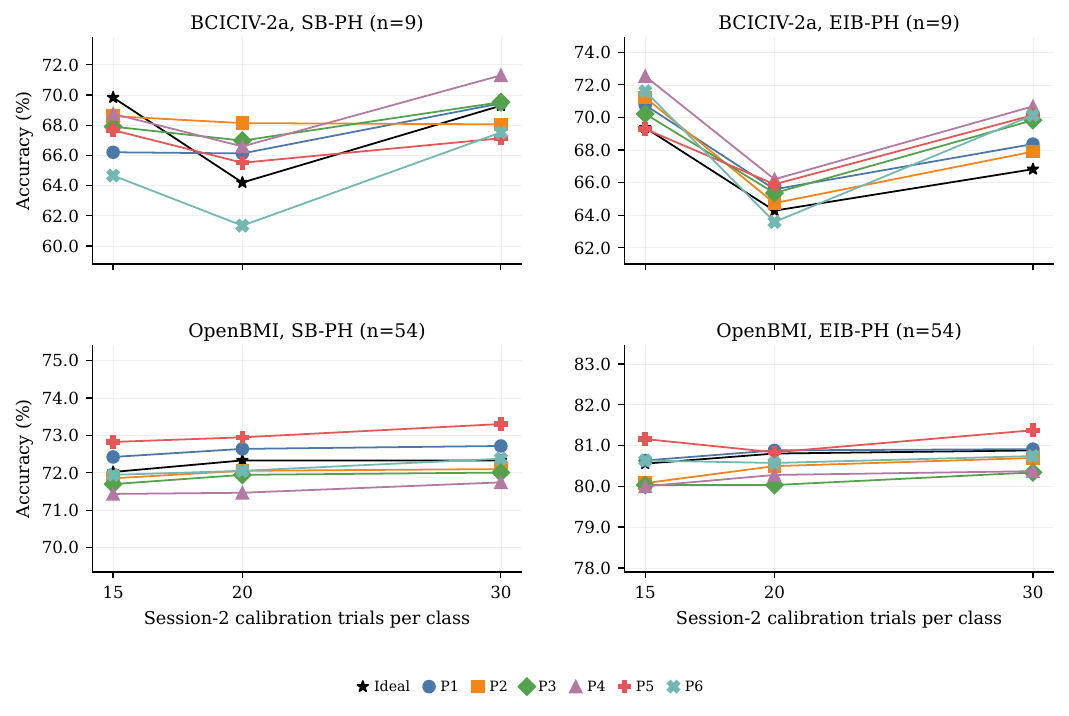}
\caption{Mean held-out decoding accuracy across Session~2 calibration budgets for the ideal-link reference and for Policies P1--P6 under the default heterogeneous-link condition. Each point denotes the cohort mean across subjects (BCICIV-2a, $n=9$; OpenBMI, $n=54$) from the primary controlled realization using seed 2026. Panels use independently scaled y-axes to show within-panel differences.}
\label{fig:supp-accuracy-calibration}
\end{figure*}

Table~\ref{tab:policy-landscape} reports the complete operating-point results for the ideal-link reference and Policies P1--P6 under the default heterogeneous-link condition. Accuracy is first averaged equally across the three Session~2 calibration budgets, $k\in\{15,20,30\}$, within each subject and is then averaged across subjects. Supplementary Figure~\ref{fig:supp-accuracy-calibration} complements these budget-averaged operating points by showing mean held-out decoding accuracy separately at each calibration budget for both datasets and learning regimes. The table and figure report the primary controlled realization using seed 2026.

Together, Table~\ref{tab:policy-landscape} and Supplementary Figure~\ref{fig:supp-accuracy-calibration} characterize each policy as a joint learning and synchronization operating point. Scheduling, buffering, stale-update admission, and backbone-download control determine which local updates reach the coordinator and which backbone versions remain available to the gateways. Consequently, policy changes alter the collaborative training trajectory rather than merely changing a communication cost measured after training. P5 and P6 produce the lowest server-to-client traffic in both datasets because they use stale-aware backbone downloading. In the primary controlled realization, P5 reduced server-to-client backbone traffic relative to P3 by 41.34\% on BCICIV-2a and 41.38\% on OpenBMI, corresponding to decreases from 6.27 to 3.68~MB and from 37.49 to 21.98~MB, respectively. The associated accuracy differences vary by dataset and regime, which motivates the matched repeated-run analysis in Section~\ref{sec:supp-repeated-runs}. The repeated analysis is restricted to P3 and P5; the P1, P2, P4, and P6 operating points therefore correspond to the primary controlled realization.

\begin{table*}[!t]
\centering
\caption{Accuracy and communication operating points under the ideal-link reference and the default heterogeneous-link condition.}
\label{tab:policy-landscape}
\begingroup
\scriptsize
\setlength{\tabcolsep}{2.5pt}
\renewcommand{\arraystretch}{1.06}

\resizebox{\textwidth}{!}{%
\begin{tabular}{@{}L{0.085\textheight}L{0.065\textheight}L{0.105\textheight}C{0.060\textheight}C{0.115\textheight}C{0.080\textheight}C{0.080\textheight}C{0.090\textheight}C{0.105\textheight}C{0.100\textheight}@{}}
\toprule
\textbf{Dataset} &
\textbf{Regime} &
\makecell{\textbf{Link}\\\textbf{condition}} &
\textbf{Policy} &
\makecell{\textbf{Subject accuracy}\\\textbf{(\%, mean $\pm$ SD)}} &
\makecell{\textbf{C2S update}\\\textbf{traffic (MB)}} &
\makecell{\textbf{S2C backbone}\\\textbf{traffic (MB)}} &
\makecell{\textbf{Coordinator-}\\\textbf{rejected uploads}\\\textbf{(\%)}} &
\makecell{\textbf{Mean accepted-}\\\textbf{update staleness}\\\textbf{(versions)}} &
\makecell{\textbf{Mean buffered-}\\\textbf{upload delay}\\\textbf{(rounds)}} \\
\midrule
BCICIV-2a & SB-PH & Ideal link & Ideal & $67.77 \pm 11.67$ & 9.30 & 9.23 & 0.00 & 0.00 & \textemdash \\
BCICIV-2a & SB-PH & Heterogeneous link & P1 & $67.26 \pm 11.17$ & 6.29 & 6.27 & 0.00 & 0.00 & \textemdash \\
BCICIV-2a & SB-PH & Heterogeneous link & P2 & $68.26 \pm 9.37$ & 9.02 & 6.27 & 0.00 & 0.97 & 1.50 \\
BCICIV-2a & SB-PH & Heterogeneous link & P3 & $68.13 \pm 11.39$ & 9.02 & 6.27 & 19.20 & 0.27 & 1.49 \\
BCICIV-2a & SB-PH & Heterogeneous link & P4 & $68.88 \pm 10.58$ & 7.98 & 6.27 & 8.86 & 0.27 & 1.00 \\
BCICIV-2a & SB-PH & Heterogeneous link & P5 & $66.77 \pm 10.29$ & 8.44 & 3.68 & 24.59 & 0.52 & 1.49 \\
BCICIV-2a & SB-PH & Heterogeneous link & P6 & $64.51 \pm 9.02$ & 7.40 & 3.68 & 14.20 & 0.52 & 1.00 \\
\addlinespace[2pt]
BCICIV-2a & EIB-PH & Ideal link & Ideal & $66.82 \pm 13.15$ & 9.30 & 9.23 & 0.00 & 0.00 & \textemdash \\
BCICIV-2a & EIB-PH & Heterogeneous link & P1 & $68.24 \pm 11.28$ & 6.29 & 6.27 & 0.00 & 0.00 & \textemdash \\
BCICIV-2a & EIB-PH & Heterogeneous link & P2 & $67.95 \pm 13.09$ & 9.02 & 6.27 & 0.00 & 0.97 & 1.50 \\
BCICIV-2a & EIB-PH & Heterogeneous link & P3 & $68.47 \pm 10.05$ & 9.02 & 6.27 & 19.20 & 0.27 & 1.49 \\
BCICIV-2a & EIB-PH & Heterogeneous link & P4 & $69.80 \pm 9.52$ & 7.98 & 6.27 & 8.86 & 0.27 & 1.00 \\
BCICIV-2a & EIB-PH & Heterogeneous link & P5 & $68.44 \pm 9.61$ & 8.44 & 3.68 & 24.59 & 0.52 & 1.49 \\
BCICIV-2a & EIB-PH & Heterogeneous link & P6 & $68.44 \pm 10.51$ & 7.40 & 3.68 & 14.20 & 0.52 & 1.00 \\
\midrule
OpenBMI & SB-PH & Ideal link & Ideal & $72.23 \pm 12.73$ & 55.00 & 54.56 & 0.00 & 0.00 & \textemdash \\
OpenBMI & SB-PH & Heterogeneous link & P1 & $72.59 \pm 12.35$ & 37.65 & 37.49 & 0.00 & 0.00 & \textemdash \\
OpenBMI & SB-PH & Heterogeneous link & P2 & $72.00 \pm 11.57$ & 52.83 & 37.49 & 0.00 & 0.97 & 1.50 \\
OpenBMI & SB-PH & Heterogeneous link & P3 & $71.88 \pm 12.00$ & 52.83 & 37.49 & 18.51 & 0.25 & 1.50 \\
OpenBMI & SB-PH & Heterogeneous link & P4 & $71.55 \pm 12.53$ & 47.00 & 37.49 & 8.45 & 0.25 & 1.00 \\
OpenBMI & SB-PH & Heterogeneous link & P5 & $73.02 \pm 12.18$ & 52.62 & 21.98 & 22.03 & 0.55 & 1.49 \\
OpenBMI & SB-PH & Heterogeneous link & P6 & $72.12 \pm 11.83$ & 46.79 & 21.98 & 12.37 & 0.55 & 1.00 \\
\addlinespace[2pt]
OpenBMI & EIB-PH & Ideal link & Ideal & $80.75 \pm 12.35$ & 55.00 & 54.56 & 0.00 & 0.00 & \textemdash \\
OpenBMI & EIB-PH & Heterogeneous link & P1 & $80.81 \pm 12.56$ & 37.65 & 37.49 & 0.00 & 0.00 & \textemdash \\
OpenBMI & EIB-PH & Heterogeneous link & P2 & $80.42 \pm 12.38$ & 52.83 & 37.49 & 0.00 & 0.97 & 1.50 \\
OpenBMI & EIB-PH & Heterogeneous link & P3 & $80.13 \pm 12.36$ & 52.83 & 37.49 & 18.51 & 0.25 & 1.50 \\
OpenBMI & EIB-PH & Heterogeneous link & P4 & $80.22 \pm 12.81$ & 47.00 & 37.49 & 8.45 & 0.25 & 1.00 \\
OpenBMI & EIB-PH & Heterogeneous link & P5 & $81.12 \pm 12.26$ & 52.62 & 21.98 & 22.03 & 0.55 & 1.49 \\
OpenBMI & EIB-PH & Heterogeneous link & P6 & $80.65 \pm 12.57$ & 46.79 & 21.98 & 12.37 & 0.55 & 1.00 \\
\bottomrule
\end{tabular}%
}

\vspace{0.4em}
\begin{minipage}{\textwidth}
\scriptsize
\emph{Notes:} For each subject, accuracy is first averaged equally across the three calibration budgets, $k\in\{15,20,30\}$; the reported accuracy is the cohort mean $\pm$ sample SD across these subject-level averages. Accuracy and rejected-upload rates are reported in percent, traffic in decimal MB, accepted-update staleness in backbone versions, and buffered-upload delay in communication rounds. C2S and S2C denote client-to-server and server-to-client, respectively; a dash indicates that no buffered-upload event occurred.
\end{minipage}

\endgroup
\end{table*}

\FloatBarrier

\section{Robustness Across Training and Gateway-Availability Realizations}
\label{sec:supp-repeated-runs}

\subsection{Experimental Design and Statistical Analysis}

The principal P3/P5 comparison was repeated under the default heterogeneous-link condition using five matched replicates for each dataset and learning regime. Across the five replicates, model seeds 2026--2030 were paired with availability-trace seeds 12026--12030 and replicate-specific tie-break seeds 22026--22030 for P5. The P5 priority-scheduling rule itself remained fixed across replicates, with these seeds used only to resolve otherwise tied gateway priorities; the gateway-group seed remained fixed at 2026. Within each replicate pair, P3 and P5 used the same initial model state, subject-by-round availability trace, gateway-group assignment, Session~2 partition, calibration samples, and stochastic state at the start of federated training. For EIB-PH, each matched P3/P5 pair also shared the same pooled Session~1 pretrained backbone. The design comprised five matched P3/P5 replicate pairs for each dataset--regime combination, yielding 40 completed policy runs in total.

For each subject and policy, accuracy was averaged equally across $k\in\{15,20,30\}$. Replicate-level cohort accuracy was then obtained by averaging across subjects, and the paired replicate effect was defined as the P5 cohort mean minus the corresponding P3 cohort mean. Sample SD, the observed range, and a Student-t 95\% confidence interval summarize variation across the five replicates for both cohort accuracy and the paired P5-minus-P3 effect. Because the number of replicate pairs is small, the Student-$t$ interval is descriptive. The primary interval for the repeated accuracy contrast is a crossed hierarchical bootstrap with 10,000 draws. Replicate pairs and subjects were resampled independently with replacement while preserving the within-subject P3/P5 pairing. Across replicates, training initialization and the realized gateway-availability trace were varied, while the P5 priority-scheduling rule remained fixed. The repeated analysis therefore evaluates robustness across these matched stochastic realizations without attributing run-to-run differences to any single source.

\begin{table*}[t]
\centering
\caption{Accuracy robustness of P3 and P5 across five matched replicates.}
\label{tab:supp-repeated-accuracy}
\begingroup
\scriptsize
\setlength{\tabcolsep}{2.5pt}
\renewcommand{\arraystretch}{1.08}

\resizebox{\textwidth}{!}{%
\begin{tabular}{@{}L{0.090\textheight}L{0.075\textheight}C{0.050\textheight}C{0.105\textheight}C{0.120\textheight}C{0.105\textheight}C{0.125\textheight}C{0.115\textheight}C{0.110\textheight}@{}}
\toprule
\multicolumn{9}{c}{\textbf{A. Replicate-level cohort accuracy}} \\
\midrule
\textbf{Dataset} &
\textbf{Regime} &
\textbf{Policy} &
\textbf{$n$} &
\makecell{\textbf{Mean subject}\\\textbf{accuracy across}\\\textbf{budgets (\%)}} &
\makecell{\textbf{Sample SD across}\\\textbf{replicates}} &
\makecell{\textbf{Student-$t$ 95\% CI}\\\textbf{(\%)}} &
\makecell{\textbf{Observed replicate}\\\textbf{range (\%)}} &
\makecell{\textbf{Number of matched}\\\textbf{replicates}} \\
\midrule
BCICIV-2a & SB-PH  & P3 & 9  & 68.00 & 1.52 & $[66.11,\ 69.88]$ & $[66.36,\ 69.83]$ & 5 \\
BCICIV-2a & SB-PH  & P5 & 9  & 68.80 & 0.99 & $[67.57,\ 70.02]$ & $[67.67,\ 69.65]$ & 5 \\
\addlinespace[2pt]
BCICIV-2a & EIB-PH & P3 & 9  & 68.61 & 1.23 & $[67.09,\ 70.13]$ & $[67.31,\ 70.63]$ & 5 \\
BCICIV-2a & EIB-PH & P5 & 9  & 69.50 & 1.50 & $[67.63,\ 71.36]$ & $[67.90,\ 71.71]$ & 5 \\
\midrule
OpenBMI & SB-PH  & P3 & 54 & 72.24 & 0.36 & $[71.80,\ 72.69]$ & $[71.68,\ 72.55]$ & 5 \\
OpenBMI & SB-PH  & P5 & 54 & 72.15 & 0.34 & $[71.72,\ 72.58]$ & $[71.66,\ 72.63]$ & 5 \\
\addlinespace[2pt]
OpenBMI & EIB-PH & P3 & 54 & 80.12 & 1.06 & $[78.80,\ 81.44]$ & $[78.52,\ 81.10]$ & 5 \\
OpenBMI & EIB-PH & P5 & 54 & 80.05 & 1.17 & $[78.59,\ 81.51]$ & $[78.25,\ 80.96]$ & 5 \\
\midrule
\multicolumn{9}{c}{\textbf{B. Paired P5-versus-P3 accuracy effects}} \\
\midrule
\textbf{Dataset} &
\textbf{Regime} &
\textbf{$n$} &
\makecell{\textbf{Mean paired}\\\textbf{P5--P3}\\\textbf{effect (pp)}} &
\makecell{\textbf{Sample SD across}\\\textbf{replicates}} &
\makecell{\textbf{Student-$t$ 95\% CI}\\\textbf{(pp)}} &
\makecell{\textbf{Crossed hierarchical}\\\textbf{95\% CI (pp)}} &
\makecell{\textbf{Observed replicate}\\\textbf{range (pp)}} &
\makecell{\textbf{Replicates with}\\\textbf{positive P5--P3}\\\textbf{effect}} \\
\midrule
BCICIV-2a & SB-PH  & 9  & $+0.80$ & 2.06 & $[-1.76,\ 3.35]$ & $[-1.04,\ 3.08]$ & $[-1.08,\ 3.14]$ & 2/5 \\
BCICIV-2a & EIB-PH & 9  & $+0.88$ & 0.95 & $[-0.29,\ 2.06]$ & $[-1.16,\ 2.97]$ & $[-0.28,\ 1.90]$ & 4/5 \\
\midrule
OpenBMI & SB-PH  & 54 & $-0.09$ & 0.40 & $[-0.59,\ 0.41]$ & $[-0.78,\ 0.67]$ & $[-0.52,\ 0.50]$ & 2/5 \\
OpenBMI & EIB-PH & 54 & $-0.07$ & 0.16 & $[-0.26,\ 0.12]$ & $[-0.53,\ 0.39]$ & $[-0.27,\ 0.11]$ & 2/5 \\
\bottomrule
\end{tabular}%
}

\vspace{0.4em}
\begin{minipage}{\textwidth}
\scriptsize
\emph{Notes:} Panel A summarizes replicate-level cohort accuracy, and Panel B reports paired P5-minus-P3 accuracy effects in percentage points. The Student-$t$ intervals summarize variation across the five matched replicates. The crossed hierarchical 95\% CIs are obtained from resampling matched replicates and subjects while preserving the within-subject P3/P5 pairing.
\end{minipage}

\endgroup
\end{table*}

\subsection{Accuracy Robustness}

Table~\ref{tab:supp-repeated-accuracy} summarizes the repeated accuracy results. On BCICIV-2a, the mean P5-minus-P3 effect is $+0.80$ percentage points under SB-PH and $+0.88$ percentage points under EIB-PH. The effect is positive in two of five SB-PH replicates and four of five EIB-PH replicates. The corresponding hierarchical 95\% confidence intervals are $[-1.04,\ 3.08]$ and $[-1.16,\ 2.97]$ percentage points. On OpenBMI, the mean effects are $-0.09$ percentage points under SB-PH and $-0.07$ percentage points under EIB-PH, with two positive replicates in each regime. Their hierarchical intervals are $[-0.78,\ 0.67]$ and $[-0.53,\ 0.39]$ percentage points. All four hierarchical intervals include zero.

In the primary controlled OpenBMI analysis, the P5-minus-P3 effects were $+1.14$ percentage points under SB-PH and $+0.99$ percentage points under EIB-PH. Across the five matched replicates, the corresponding effects ranged from $-0.52$ to $+0.50$ percentage points and from $-0.27$ to $+0.11$ percentage points, respectively. Thus, the positive effects observed in the primary analysis were not reproduced consistently across the repeated robustness conditions.

\subsection{Subject-Level Heterogeneity}

Table~\ref{tab:supp-repeated-subject-effects} summarizes subject-level P5-minus-P3 effects across the five replicates. Positive and negative effects occur in every dataset--regime condition. On BCICIV-2a, the subject-mean effects range from $-0.93$ to $+3.33$ percentage points under SB-PH and from $-2.96$ to $+6.02$ percentage points under EIB-PH. The corresponding OpenBMI ranges are $-4.22$ to $+4.11$ percentage points and $-3.78$ to $+3.17$ percentage points. The full subject-by-replicate ranges are wider, reaching $[-6.94,\ 10.19]$ percentage points for BCICIV-2a SB-PH and $[-12.50,\ 21.94]$ percentage points for OpenBMI SB-PH. Subject-by-replicate effects of at least five percentage points in both the positive and negative directions were observed under SB-PH and EIB-PH on both datasets. Similarity in cohort means therefore does not imply uniform subject-level behavior across synchronization policies.

\begin{table*}[t]
\centering
\caption{Subject-level P5/P3 accuracy effects across five matched replicates.}
\label{tab:supp-repeated-subject-effects}
\begingroup
\scriptsize
\setlength{\tabcolsep}{2.5pt}
\renewcommand{\arraystretch}{1.08}

\resizebox{\textwidth}{!}{%
\begin{tabular}{@{}L{0.095\textheight}L{0.075\textheight}C{0.045\textheight}C{0.125\textheight}C{0.125\textheight}C{0.120\textheight}C{0.135\textheight}C{0.080\textheight}C{0.080\textheight}@{}}
\toprule
\textbf{Dataset} &
\textbf{Regime} &
\textbf{$n$} &
\makecell{\textbf{Median subject-mean}\\\textbf{P5--P3 effect (pp)}} &
\makecell{\textbf{Observed}\\\textbf{subject-mean}\\\textbf{range (pp)}} &
\makecell{\textbf{Subjects with positive}\\\textbf{mean effect}} &
\makecell{\textbf{Full subject-by-replicate}\\\textbf{range (pp)}} &
\makecell{\textbf{$\leq-5$ pp}\\\textbf{effects}} &
\makecell{\textbf{$\geq+5$ pp}\\\textbf{effects}} \\
\midrule
BCICIV-2a & SB-PH  & 9  & $+0.83$ & $[-0.93,\ 3.33]$ & 6/9  & $[-6.94,\ 10.19]$ & 4/45  & 6/45 \\
BCICIV-2a & EIB-PH & 9  & $+0.88$ & $[-2.96,\ 6.02]$ & 6/9  & $[-7.64,\ 9.03]$  & 5/45  & 9/45 \\
\midrule
OpenBMI & SB-PH  & 54 & $-0.08$ & $[-4.22,\ 4.11]$ & 25/54 & $[-12.50,\ 21.94]$ & 29/270 & 25/270 \\
OpenBMI & EIB-PH & 54 & $-0.22$ & $[-3.78,\ 3.17]$ & 21/54 & $[-10.56,\ 7.50]$  & 9/270  & 11/270 \\
\bottomrule
\end{tabular}%
}

\vspace{0.4em}
\begin{minipage}{\textwidth}
\scriptsize
\emph{Notes:} Subject-mean effects average each subject's paired P5-minus-P3 accuracy effect across the five replicates; positive-subject counts include subjects with a strictly positive mean effect. The full subject-by-replicate range and the $\pm5$-pp counts are calculated over all matched subject-by-replicate effects, yielding 45 comparisons for BCICIV-2a and 270 for OpenBMI. Effects exactly equal to $-5$ or $+5$~pp are included in the corresponding threshold counts.
\end{minipage}

\endgroup
\end{table*}

\FloatBarrier

\subsection{Communication Robustness}

Table~\ref{tab:supp-repeated-communication} reports communication outcomes across the five matched replicates. The reported communication values are common to SB-PH and EIB-PH at the displayed precision. P5 reduces server-to-client backbone traffic in every replicate. The mean reduction is $42.23\%$ on BCICIV-2a, with an observed range of $41.59\%$ to $42.76\%$, and $41.98\%$ on OpenBMI, with a range of $41.47\%$ to $42.47\%$.

P5 also reduces total communication in all five replicates. Relative to P3, mean total traffic decreases by $3.20$~MB on BCICIV-2a and $15.91$~MB on OpenBMI. These reductions are accompanied by increases in the coordinator-rejected upload rate of $5.29$ and $3.91$ percentage points, respectively, and by increases in mean accepted-update staleness of $0.266$ and $0.299$ backbone versions. Mean buffered-upload delay differs by no more than 0.003 rounds on either dataset. The communication savings therefore arise primarily from selective synchronization and avoided backbone downloads rather than from longer buffering intervals.

\begin{table*}[t]
\centering
\caption{Communication robustness of P3 and P5 across five matched replicates.}
\label{tab:supp-repeated-communication}
\begingroup
\scriptsize
\setlength{\tabcolsep}{2.5pt}
\renewcommand{\arraystretch}{1.06}

\resizebox{\textwidth}{!}{%
\begin{tabular}{@{}L{0.090\textheight}L{0.135\textheight}C{0.050\textheight}C{0.100\textheight}C{0.100\textheight}C{0.120\textheight}C{0.115\textheight}C{0.105\textheight}C{0.095\textheight}@{}}
\toprule
\textbf{Dataset} &
\textbf{Metric} &
\textbf{Unit} &
\makecell{\textbf{P3}\\\textbf{mean $\pm$ SD}} &
\makecell{\textbf{P5}\\\textbf{mean $\pm$ SD}} &
\makecell{\textbf{Paired difference}\\\textbf{mean $\pm$ SD}} &
\makecell{\textbf{Student-$t$}\\\textbf{95\% CI}} &
\makecell{\textbf{Observed}\\\textbf{range}} &
\makecell{\textbf{Direction count}\\\textbf{($+/-/=$)}} \\
\midrule
BCICIV-2a & C2S traffic & MB & $8.765 \pm 0.082$ & $8.215 \pm 0.101$ & $-0.550 \pm 0.043$ & $[-0.603,\ -0.496]$ & $[-0.603,\ -0.482]$ & 0/5/0 \\
BCICIV-2a & S2C traffic & MB & $6.279 \pm 0.055$ & $3.628 \pm 0.017$ & $-2.652 \pm 0.054$ & $[-2.718,\ -2.585]$ & $[-2.710,\ -2.599]$ & 0/5/0 \\
BCICIV-2a & Total traffic & MB & $15.044 \pm 0.066$ & $11.843 \pm 0.107$ & $-3.201 \pm 0.091$ & $[-3.314,\ -3.088]$ & $[-3.313,\ -3.081]$ & 0/5/0 \\
BCICIV-2a & Coordinator-rejected uploads & pp & $17.962 \pm 1.050$ & $23.251 \pm 0.956$ & $+5.290 \pm 0.294$ & $[4.925,\ 5.655]$ & $[4.817,\ 5.579]$ & 5/0/0 \\
BCICIV-2a & Accepted-update staleness & versions & $0.244 \pm 0.010$ & $0.510 \pm 0.008$ & $+0.266 \pm 0.007$ & $[0.257,\ 0.274]$ & $[0.255,\ 0.272]$ & 5/0/0 \\
BCICIV-2a & Buffered-upload delay & rounds & $1.489 \pm 0.018$ & $1.487 \pm 0.019$ & $-0.002 \pm 0.003$ & $[-0.006,\ 0.002]$ & $[-0.006,\ 0.000]$ & 0/2/3 \\
BCICIV-2a & Avoided S2C downloads & pp & $0.000 \pm 0.000$ & $37.326 \pm 0.424$ & $+37.326 \pm 0.424$ & $[36.799,\ 37.852]$ & $[36.667,\ 37.744]$ & 5/0/0 \\
BCICIV-2a & S2C traffic reduction & \% & \textemdash & \textemdash & $+42.226 \pm 0.515$ & $[41.587,\ 42.865]$ & $[41.589,\ 42.762]$ & 5/0/0 \\
\midrule
OpenBMI & C2S traffic & MB & $52.211 \pm 0.195$ & $51.994 \pm 0.153$ & $-0.217 \pm 0.050$ & $[-0.280,\ -0.155]$ & $[-0.272,\ -0.136]$ & 0/5/0 \\
OpenBMI & S2C traffic & MB & $37.392 \pm 0.417$ & $21.696 \pm 0.148$ & $-15.697 \pm 0.303$ & $[-16.074,\ -15.320]$ & $[-16.083,\ -15.401]$ & 0/5/0 \\
OpenBMI & Total traffic & MB & $89.604 \pm 0.606$ & $73.690 \pm 0.290$ & $-15.914 \pm 0.345$ & $[-16.343,\ -15.486]$ & $[-16.354,\ -15.536]$ & 0/5/0 \\
OpenBMI & Coordinator-rejected uploads & pp & $17.810 \pm 0.524$ & $21.716 \pm 0.406$ & $+3.906 \pm 0.282$ & $[3.557,\ 4.256]$ & $[3.603,\ 4.275]$ & 5/0/0 \\
OpenBMI & Accepted-update staleness & versions & $0.249 \pm 0.006$ & $0.548 \pm 0.004$ & $+0.299 \pm 0.004$ & $[0.294,\ 0.303]$ & $[0.294,\ 0.304]$ & 5/0/0 \\
OpenBMI & Buffered-upload delay & rounds & $1.489 \pm 0.012$ & $1.487 \pm 0.011$ & $-0.003 \pm 0.002$ & $[-0.005,\ -0.001]$ & $[-0.005,\ 0.000]$ & 0/4/1 \\
OpenBMI & Avoided S2C downloads & pp & $0.000 \pm 0.000$ & $42.154 \pm 0.345$ & $+42.154 \pm 0.345$ & $[41.725,\ 42.583]$ & $[41.644,\ 42.566]$ & 5/0/0 \\
OpenBMI & S2C traffic reduction & \% & \textemdash & \textemdash & $+41.976 \pm 0.380$ & $[41.505,\ 42.447]$ & $[41.466,\ 42.465]$ & 5/0/0 \\
\bottomrule
\end{tabular}%
}

\vspace{0.4em}
\begin{minipage}{\textwidth}
\scriptsize
\emph{Notes:} P3 and P5 columns report replicate-level mean $\pm$ sample SD. Paired differences are computed as P5 minus P3, except S2C traffic reduction, which is measured relative to P3 and is positive when P5 uses less S2C traffic. Direction counts $(+/-/=)$ give the number of matched replicates with positive, negative, or zero P5-minus-P3 differences; for S2C traffic reduction, a positive direction denotes lower S2C traffic under P5.
\end{minipage}

\endgroup
\end{table*}

\subsection{Replicate-Level Results and Interpretation}

Table~\ref{tab:supp-complete-replicates} reports the complete replicate-level cohort accuracies and server-to-client reductions. On BCICIV-2a, the accuracy effect ranges from $-1.08$ to $+3.14$ percentage points under SB-PH and from $-0.28$ to $+1.90$ percentage points under EIB-PH. On OpenBMI, the corresponding ranges are $-0.52$ to $+0.50$ percentage points and $-0.27$ to $+0.11$ percentage points. By contrast, the server-to-client reduction remains close to $42\%$ in every replicate.

Across the five matched replicates, P5 consistently reduced server-to-client backbone traffic and total communication. The accuracy differences were comparatively small, varied in direction across replicates, and had crossed hierarchical-bootstrap 95\% confidence intervals that included zero in all four dataset--regime conditions. Taken together, these results show that the communication reduction is consistent across the evaluated replicates, whereas the accuracy effect remains realization-dependent.

\begin{table*}[t]
\centering
\caption{Complete replicate-level P3/P5 results across five matched replicates.}
\label{tab:supp-complete-replicates}
\begingroup
\scriptsize
\setlength{\tabcolsep}{2.5pt}
\renewcommand{\arraystretch}{1.06}

\resizebox{\textwidth}{!}{%
\begin{tabular}{@{}L{0.100\textheight}L{0.075\textheight}C{0.065\textheight}C{0.090\textheight}C{0.090\textheight}C{0.105\textheight}C{0.105\textheight}C{0.110\textheight}C{0.110\textheight}@{}}
\toprule
\textbf{Dataset} &
\textbf{Regime} &
\makecell{\textbf{Model}\\\textbf{seed}} &
\makecell{\textbf{Availability-}\\\textbf{trace seed}} &
\makecell{\textbf{P5 scheduler} \\\textbf{tie-break seed}} &
\makecell{\textbf{P3 mean subject}\\\textbf{accuracy (\%)}} &
\makecell{\textbf{P5 mean subject}\\\textbf{accuracy (\%)}} &
\makecell{\textbf{P5--P3 accuracy}\\\textbf{difference (pp)}} &
\makecell{\textbf{S2C traffic}\\\textbf{reduction (\%)}} \\
\midrule
BCICIV-2a & SB-PH  & 2026 & 12026 & 22026 & 66.36 & 69.50 & $+3.14$ & 41.82 \\
BCICIV-2a & SB-PH  & 2027 & 12027 & 22027 & 68.57 & 67.77 & $-0.80$ & 42.76 \\
BCICIV-2a & SB-PH  & 2028 & 12028 & 22028 & 69.83 & 69.65 & $-0.18$ & 42.29 \\
BCICIV-2a & SB-PH  & 2029 & 12029 & 22029 & 68.75 & 67.67 & $-1.08$ & 42.67 \\
BCICIV-2a & SB-PH  & 2030 & 12030 & 22030 & 66.49 & 69.39 & $+2.91$ & 41.59 \\
\addlinespace[2pt]
BCICIV-2a & EIB-PH & 2026 & 12026 & 22026 & 68.36 & 70.27 & $+1.90$ & 41.82 \\
BCICIV-2a & EIB-PH & 2027 & 12027 & 22027 & 68.57 & 68.67 & $+0.10$ & 42.76 \\
BCICIV-2a & EIB-PH & 2028 & 12028 & 22028 & 70.63 & 71.71 & $+1.08$ & 42.29 \\
BCICIV-2a & EIB-PH & 2029 & 12029 & 22029 & 67.31 & 68.93 & $+1.62$ & 42.67 \\
BCICIV-2a & EIB-PH & 2030 & 12030 & 22030 & 68.18 & 67.90 & $-0.28$ & 41.59 \\
\midrule
OpenBMI & SB-PH  & 2026 & 12026 & 22026 & 72.55 & 72.63 & $+0.09$ & 42.47 \\
OpenBMI & SB-PH  & 2027 & 12027 & 22027 & 71.68 & 72.18 & $+0.50$ & 41.81 \\
OpenBMI & SB-PH  & 2028 & 12028 & 22028 & 72.55 & 72.17 & $-0.38$ & 42.20 \\
OpenBMI & SB-PH  & 2029 & 12029 & 22029 & 72.27 & 72.12 & $-0.15$ & 41.47 \\
OpenBMI & SB-PH  & 2030 & 12030 & 22030 & 72.18 & 71.66 & $-0.52$ & 41.95 \\
\addlinespace[2pt]
OpenBMI & EIB-PH & 2026 & 12026 & 22026 & 80.81 & 80.86 & $+0.06$ & 42.47 \\
OpenBMI & EIB-PH & 2027 & 12027 & 22027 & 79.58 & 79.47 & $-0.11$ & 41.81 \\
OpenBMI & EIB-PH & 2028 & 12028 & 22028 & 80.61 & 80.72 & $+0.11$ & 42.20 \\
OpenBMI & EIB-PH & 2029 & 12029 & 22029 & 81.10 & 80.96 & $-0.14$ & 41.47 \\
OpenBMI & EIB-PH & 2030 & 12030 & 22030 & 78.52 & 78.25 & $-0.27$ & 41.95 \\
\bottomrule
\end{tabular}%
}

\vspace{0.4em}
\begin{minipage}{\textwidth}
\scriptsize
\emph{Notes:} Cohort accuracy first averages each subject equally across the three calibration budgets and then averages across subjects. Accuracy differences are computed as P5 minus P3, while S2C traffic reductions are measured relative to P3. The gateway-group seed is fixed at 2026 across all replicates.
\end{minipage}

\endgroup
\end{table*}

\FloatBarrier

\section{Link-Availability Sensitivity}
\label{sec:supp-link-sensitivity}

Table~\ref{tab:link-sensitivity-full} reports the EIB-PH results for P3 and P5 under mild, default, and severe heterogeneous-link profiles. The online probabilities assigned to the high-, moderate-, and low-availability gateway groups are $0.98/0.85/0.60$, $0.95/0.70/0.40$, and $0.90/0.50/0.20$, respectively. P3 and P5 retain the same FIFO buffering and stale-update admission rules in all three profiles. Their comparison therefore continues to isolate the combined effect of communication-aware scheduling and stale-aware backbone downloading.

As availability decreases, both policies encounter fewer synchronization opportunities and a larger fraction of transmitted uploads is rejected by the coordinator. From the mild to the severe profile, the rejected-upload rate rises from $9.19\%$ to $28.25\%$ under P3 and from $15.39\%$ to $32.20\%$ under P5 on BCICIV-2a. The corresponding OpenBMI rates rise from $8.63\%$ to $27.52\%$ and from $13.60\%$ to $30.64\%$. Mean buffered-upload delay reaches approximately $1.6$ rounds in the severe profile, and mean accepted-update staleness remains higher under P5. Despite these changes, P5 uses less server-to-client traffic at every severity level. Its traffic is $4.07$, $3.68$, and $2.99$~MB on BCICIV-2a, compared with $7.46$, $6.27$, and $4.86$~MB under P3. On OpenBMI, the corresponding P5 values are $24.39$, $21.98$, and $17.81$~MB, compared with $44.36$, $37.49$, and $28.99$~MB under P3.

The accuracy contrast does not follow a common direction across profiles. On BCICIV-2a, the P5-minus-P3 differences are $+1.77$, $-0.03$, and $+0.16$ percentage points under mild, default, and severe availability. On OpenBMI, the corresponding differences are $+0.08$, $+0.99$, and $-0.46$ percentage points. The severity analysis therefore shows a persistent communication reduction but no profile-invariant accuracy advantage. This sensitivity analysis should be distinguished from the repeated analysis in Section~\ref{sec:supp-repeated-runs}: here, availability severity is varied within the primary controlled realization, whereas the repeated analysis evaluates five matched P3/P5 replicates under the default availability profile with different training initializations and realized gateway-availability traces, while retaining the same P5 priority-scheduling rule with replicate-specific seeded tie resolution.

\begin{table*}[t]
\centering
\caption{Sensitivity of EIB-PH to heterogeneous gateway availability.}
\label{tab:link-sensitivity-full}
\begingroup
\scriptsize
\setlength{\tabcolsep}{2.5pt}
\renewcommand{\arraystretch}{1.08}

\resizebox{\textwidth}{!}{%
\begin{tabular}{@{}L{0.095\textheight}L{0.165\textheight}C{0.055\textheight}C{0.105\textheight}C{0.090\textheight}C{0.090\textheight}C{0.115\textheight}C{0.115\textheight}C{0.105\textheight}@{}}
\toprule
\textbf{Dataset} &
\makecell{\textbf{Availability profile}\\\textbf{(high/moderate/low)}} &
\textbf{Policy} &
\makecell{\textbf{Mean subject}\\\textbf{accuracy (\%)}} &
\makecell{\textbf{C2S update}\\\textbf{traffic (MB)}} &
\makecell{\textbf{S2C backbone}\\\textbf{traffic (MB)}} &
\makecell{\textbf{Coordinator-rejected}\\\textbf{uploads (\%)}} &
\makecell{\textbf{Mean accepted-}\\\textbf{update staleness}\\\textbf{(versions)}} &
\makecell{\textbf{Mean buffered-}\\\textbf{upload delay}\\\textbf{(rounds)}} \\
\midrule
BCICIV-2a & Mild (0.98/0.85/0.60) &
P3 & 68.29 & 9.18 & 7.46 & 9.19 & 0.20 & 1.35 \\
BCICIV-2a & Mild (0.98/0.85/0.60) &
P5 & 70.06 & 7.78 & 4.07 & 15.39 & 0.45 & 1.35 \\
\addlinespace[2pt]
BCICIV-2a & Default (0.95/0.70/0.40) &
P3 & 68.47 & 9.02 & 6.27 & 19.20 & 0.27 & 1.49 \\
BCICIV-2a & Default (0.95/0.70/0.40) &
P5 & 68.44 & 8.44 & 3.68 & 24.59 & 0.52 & 1.49 \\
\addlinespace[2pt]
BCICIV-2a & Severe (0.90/0.50/0.20) &
P3 & 67.90 & 7.96 & 4.86 & 28.25 & 0.29 & 1.62 \\
BCICIV-2a & Severe (0.90/0.50/0.20) &
P5 & 68.06 & 7.84 & 2.99 & 32.20 & 0.54 & 1.60 \\
\midrule
OpenBMI & Mild (0.98/0.85/0.60) &
P3 & 80.53 & 54.66 & 44.36 & 8.63 & 0.21 & 1.33 \\
OpenBMI & Mild (0.98/0.85/0.60) &
P5 & 80.61 & 50.73 & 24.39 & 13.60 & 0.51 & 1.33 \\
\addlinespace[2pt]
OpenBMI & Default (0.95/0.70/0.40) &
P3 & 80.13 & 52.83 & 37.49 & 18.51 & 0.25 & 1.50 \\
OpenBMI & Default (0.95/0.70/0.40) &
P5 & 81.12 & 52.62 & 21.98 & 22.03 & 0.55 & 1.49 \\
\addlinespace[2pt]
OpenBMI & Severe (0.90/0.50/0.20) &
P3 & 81.19 & 45.95 & 28.99 & 27.52 & 0.26 & 1.60 \\
OpenBMI & Severe (0.90/0.50/0.20) &
P5 & 80.73 & 45.95 & 17.81 & 30.64 & 0.53 & 1.59 \\
\bottomrule
\end{tabular}%
}

\vspace{0.4em}
\begin{minipage}{\textwidth}
\scriptsize
\emph{Notes:} Availability profiles report the high/moderate/low gateway online probabilities. Mean accepted-update staleness is averaged across accepted uploads, while mean buffered-upload delay is averaged across buffered uploads that reach stale-update admission.
\end{minipage}

\endgroup
\end{table*}

\FloatBarrier

\section{Associations With Subject-Level Accuracy Differences}
\label{sec:supp-subject-associations}

Table~\ref{tab:subject-vulnerability} reports descriptive Spearman rank correlations between each subject's EIB-PH accuracy difference from the matched ideal-link reference and six candidate quantities: ideal-link accuracy, assigned gateway online probability, coordinator-rejected uploads, mean accepted-update staleness, updates placed in the buffer, and avoided backbone downloads. Uncertainty is summarized using 95\% percentile confidence intervals from 10,000 subject-level bootstrap resamples within each dataset--policy analysis. These analyses characterize monotonic associations within the primary controlled realization and do not establish causal relationships or isolate the contribution of any individual communication quantity.

\begin{table*}[!b]
\centering
\caption{Associations of baseline accuracy and communication exposure with subject-level accuracy differences under EIB-PH.}
\label{tab:subject-vulnerability}
\begingroup
\scriptsize
\setlength{\tabcolsep}{2.5pt}
\renewcommand{\arraystretch}{1.08}
\resizebox{\textwidth}{!}{%
\begin{tabular}{@{}L{0.095\textheight}C{0.055\textheight}C{0.045\textheight}C{0.110\textheight}C{0.125\textheight}C{0.125\textheight}C{0.125\textheight}C{0.115\textheight}C{0.115\textheight}@{}}
\toprule
\multicolumn{3}{c}{} &
\multicolumn{6}{c}{\textbf{Spearman correlation with subject-level accuracy difference ($\rho$; 95\% CI)}} \\
\cmidrule(lr){4-9}
\textbf{Dataset} &
\textbf{Policy} &
\textbf{$n$} &
\makecell{\textbf{Ideal-link}\\\textbf{accuracy}} &
\makecell{\textbf{Assigned gateway}\\\textbf{online probability}} &
\makecell{\textbf{Coordinator-rejected}\\\textbf{uploads (count)}} &
\makecell{\textbf{Mean accepted-update}\\\textbf{staleness (versions)}} &
\makecell{\textbf{Updates placed}\\\textbf{in buffer (count)}} &
\makecell{\textbf{Avoided backbone}\\\textbf{downloads (count)}} \\
\midrule
BCICIV-2a & P3 & 9 &
\makecell{$-0.68$\\$[-1.00,\,0.11]$} &
\makecell{$0.26$\\$[-0.57,\,0.83]$} &
\makecell{$-0.08$\\$[-0.74,\,0.74]$} &
\makecell{$-0.25$\\$[-0.83,\,0.64]$} &
\makecell{$-0.14$\\$[-0.83,\,0.74]$} &
\textemdash \\
BCICIV-2a & P5 & 9 &
\makecell{$-0.68$\\$[-1.00,\,0.07]$} &
\makecell{$0.21$\\$[-0.59,\,0.82]$} &
\makecell{$0.03$\\$[-0.72,\,0.73]$} &
\makecell{$0.13$\\$[-0.71,\,0.72]$} &
\makecell{$0.00$\\$[-0.75,\,0.71]$} &
\makecell{$0.05$\\$[-0.64,\,0.83]$} \\
\midrule
OpenBMI & P3 & 54 &
\makecell{$-0.17$\\$[-0.42,\,0.10]$} &
\makecell{$0.09$\\$[-0.19,\,0.37]$} &
\makecell{$-0.07$\\$[-0.35,\,0.21]$} &
\makecell{$-0.10$\\$[-0.38,\,0.19]$} &
\makecell{$-0.08$\\$[-0.36,\,0.20]$} &
\textemdash \\
OpenBMI & P5 & 54 &
\makecell{$-0.14$\\$[-0.42,\,0.16]$} &
\makecell{$0.05$\\$[-0.23,\,0.31]$} &
\makecell{$0.03$\\$[-0.24,\,0.32]$} &
\makecell{$-0.11$\\$[-0.36,\,0.16]$} &
\makecell{$-0.02$\\$[-0.28,\,0.24]$} &
\makecell{$0.03$\\$[-0.23,\,0.29]$} \\
\bottomrule
\end{tabular}%
}

\vspace{0.4em}
\begin{minipage}{\textwidth}
\scriptsize
\emph{Notes:} Entries report Spearman rank correlations with each policy's subject-level accuracy difference from the matched ideal-link reference, with 95\% percentile CIs obtained from 10,000 subject-level bootstrap resamples (seed 2026) within each dataset--policy analysis. The CIs are not adjusted for multiple comparisons, and the associations are descriptive rather than causal. A dash indicates that the correlation is undefined because the corresponding quantity does not vary across subjects.
\end{minipage}

\endgroup
\end{table*}

\FloatBarrier

The largest point estimates occur on BCICIV-2a, where ideal-link accuracy has a Spearman correlation of $\rho=-0.68$ with the subject-level accuracy difference under both P3 and P5. However, the corresponding 95\% bootstrap confidence intervals are wide and include zero: $[-1.00,\,0.11]$ for P3 and $[-1.00,\,0.07]$ for P5. The remaining BCICIV-2a associations are smaller in magnitude and likewise have confidence intervals that include zero. On OpenBMI, all point estimates are small, ranging from $-0.17$ to $0.09$, and all corresponding confidence intervals also include zero. The avoided-download correlation is undefined for P3 because this quantity does not vary across subjects under the P3 download rule.

Across the evaluated datasets and policies, no recorded communication-exposure quantity shows a consistent association with subject-level accuracy differences. In particular, the present correlations do not support attributing subject-level degradation to rejected uploads, accepted-update staleness, buffering, gateway availability, or avoided downloads individually. The results should therefore be interpreted as exploratory descriptions of subject-level covariation rather than as evidence of causal mechanisms or reliable predictors of vulnerability. This caution is especially important for BCICIV-2a because the analysis includes only nine subjects, and the reported confidence intervals are not adjusted for the multiple correlations examined.

\end{document}